\documentclass[journal]{IEEEtran}
\usepackage{amsmath,amsfonts}
\usepackage{algorithmic}
\usepackage{algorithm}
\usepackage{array}
\usepackage{multirow}
\usepackage[caption=false,font=normalsize,labelfont=sf,textfont=sf]{subfig}
\usepackage{textcomp}
\usepackage{stfloats}
\usepackage{makecell}
\usepackage{url}
\usepackage{verbatim}
\usepackage{graphicx}
\usepackage{cite}
\newtheorem{remark}{Remark}

\begin{document}

\title{From Distance to Direction: Structure-aware Label-specific\\ Feature Fusion for Label Distribution Learning}

\author{Suping Xu, Chuyi Dai, Lin Shang,~\IEEEmembership{Member,~IEEE,} \\Changbin Shao, Xibei Yang, and Witold Pedrycz,~\IEEEmembership{Life Fellow,~IEEE}

\thanks{S. Xu and C. Dai are with the Department of Electrical and Computer Engineering, University of Alberta, Edmonton, AB T6G 2R3, Canada (E-mail: supingxu@yahoo.com; suping2@ualberta.ca, cdai4@ualberta.ca).}
\thanks{L. Shang is with the School of Computer Science, Nanjing University, Nanjing 210023, China, and also with the State Key Laboratory for Novel Software Technology, Nanjing University, Nanjing 210023, China (E-mail: shanglin@nju.edu.cn).}
\thanks{C. Shao and X. Yang are with the School of Computer, Jiangsu University of Science and Technology, Zhenjiang 212003, China (E-mail: shaocb@just.edu.cn, jsjxy\_yxb@just.edu.cn).}
\thanks{W. Pedrycz is with the Department of Electrical and Computer Engineering, University of Alberta, Edmonton, AB T6G 2R3, Canada, also with the Institute of Systems Engineering, Macau University of Science and Technology, Taipa 999078, China, and also with the Research Center of Performance and Productivity Analysis, Istinye University, Istanbul 34396, Türkiye (E-mail: wpedrycz@ualberta.ca).}

\thanks{Manuscript received X X, 2025; revised X X, 2025.}}

\markboth{Journal of \LaTeX\ Class Files,~Vol.~X, No.~X, August~2025}%
{Shell \MakeLowercase{\textit{et al.}}: A Sample Article Using IEEEtran.cls for IEEE Journals}


\maketitle

\begin{abstract}
Label distribution learning (LDL) is an emerging learning paradigm designed to capture the relative importance of labels for each instance. Label-specific features (LSFs), constructed by LIFT, have proven effective for learning tasks with label ambiguity by leveraging clustering-based prototypes for each label to re-characterize instances. However, directly introducing LIFT into LDL tasks can be suboptimal, as the prototypes it collects primarily reflect intra-cluster relationships while neglecting cross-cluster interactions. Additionally, constructing LSFs using multi-perspective information, rather than relying solely on Euclidean distance, provides a more robust and comprehensive representation of instances, mitigating noise and bias that may arise from a single distance perspective. To address these limitations, we introduce Structural Anchor Points (SAPs) to capture inter-cluster interactions. This leads to a novel LSFs construction strategy, LIFT-SAP, which enhances LIFT by integrating both distance and directional information of each instance relative to SAPs. Furthermore, we propose a novel LDL algorithm, Label Distribution Learning via Label-specifIc FeaTure with SAPs (LDL-LIFT-SAP), which unifies multiple label description degrees predicted from different LSF spaces into a cohesive label distribution. Extensive experiments on $15$ real-world datasets demonstrate the effectiveness of LIFT-SAP over LIFT, as well as the superiority of LDL-LIFT-SAP compared to seven other well-established algorithms.
\end{abstract}


\section{Introduction}
\label{sec:introduction}
\begin{figure}[!t]
\centering
\subfloat[Landscape photo]{\includegraphics[height=1.5in]{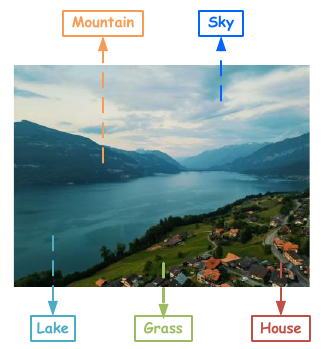}%
\label{fig:1a}}
\hfil
\subfloat[Label distribution]{\includegraphics[height=1.5in]{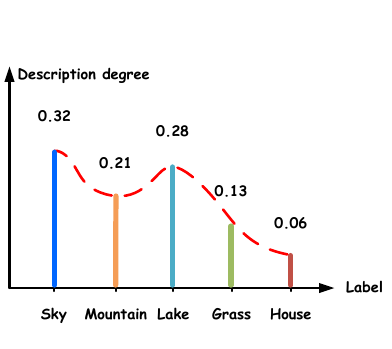}%
\label{fig:1b}}
\caption{An illustration of label distribution. (a) a landscape photo captured in Krattigen, Switzerland, containing five labels, i.e., `Sky', `Mountain', `Lake', `Grass', and `House'. (b) The corresponding label distribution of the landscape photo, with description degrees of $0.32$, $0.21$, $0.28$, $0.13$, and $0.06$ for the above labels, respectively.}
\label{fig:1}
\end{figure}

\IEEEPARstart{L}{abel} Distribution Learning (LDL)~\cite{geng:LDL} is an emerging learning paradigm that extends beyond traditional single-label learning (SLL) and multi-label learning (MLL) by assigning each instance a label distribution. Each element of the label distribution, known as the description degree of a label, explicitly quantifies the degree to which the label describes the instance. For example, Fig.\ref{fig:1} illustrates the label distribution of a landscape photo, where labels such as `Sky', `Mountain', `Lake', `Grass', and `House' contribute unequally to the description of the landscape photo. Compared to SLL and MLL, LDL offers a more nuanced representation in label space, instead of discrete binary labels. LDL has found successful applications in many real-world tasks, including facial age estimation~\cite{geng-et-al:facial-age-estimation}, facial expression recognition~\cite{lang-et-al:facial-expression}, head pose estimation~\cite{geng-et-al:head-pose-estimation}, person re-identification~\cite{qi-et-al:person-re-identification}, autism spectrum disorder classification~\cite{wang-et-al:ASD}, and pulmonary diseases prediction~\cite{wang-et-al:pulmonary-diseases-prediction}, among others.

\begin{figure}[!t]
\centering
\includegraphics[height=1.25in]{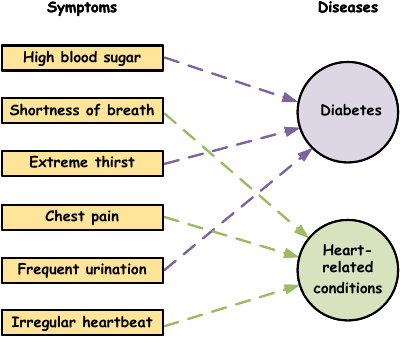}
\caption{An illustration of label-specific features}
\label{fig:LSFs}
\end{figure}

Researchers have proposed various LDL algorithms. Initially, label distribution was introduced in facial age estimation~\cite{geng-et-al:facial-age-estimation}, demonstrating its capacity to capture gradual transitions in facial appearance over time. Later, LDL was formally formalized as a learning paradigm~\cite{geng:LDL} with three principal strategies: (1) Problem Transformation (PT), which converts LDL tasks to traditional SLL tasks (e.g., PT-SVM and PT-Bayes~\cite{geng:LDL}); (2) Algorithm Adaptation (AA), which adapts existing SLL/MLL algorithms to directly handle LDL tasks (e.g., AA-$k$NN and AA-BP~\cite{geng:LDL}); and (3) Specialized Algorithm (SA), which designs dedicated algorithms for specific scenarios, effectively leveraging domain-specific insights. (e.g., the representative SA-based LDL algorithms, SA-IIS and SA-BFGS~\cite{geng:LDL}, as well as more advanced LDL algorithms reviewed in Section \ref{sec:relatedwork}). However, most existing algorithms overlook the exploration of specific semantic relations between instances and labels, as they rely on identical features across all labels. By selecting or extracting features most relevant to each label, rather than using shared original features, a more discriminative description of each label's unique characteristics is achieved, typically improving the LDL model's predictive performance by reducing noise from unrelated features. For example, as seen in Fig.\ref{fig:LSFs}, in medical diagnosis, specific symptoms such as `High blood sugar', `Frequent urination', and `Extreme thirst' are informative for discriminating diabetes, while symptoms like `Chest pain', `Shortness of breath', and `Irregular heartbeat' strongly indicate heart-related conditions. Generally, features selected or extracted to discriminate a specific label are referred to as \emph{label-specific features} (LSFs).

In fact, LSFs have garnered widespread attention, leading to the development of numerous construction strategies for MLL tasks, partial label learning (PLL) tasks, and partial multi-label learning (PML) tasks. Among these, LIFT~\cite{zhang-wu:lift} stands out as a most representative work, gathering prototypes for each label by independently performing clustering analysis on positive and negative instances, and subsequently re-characterizing each instance by querying its distances to all prototypes. Recently, LIFT has been extended into UCL~\cite{dong-et-al:label-specific}, a LSFs construction strategy in PLL tasks. UCL first employs graph-based label enhancement to yield smooth pseudo-labels and defines an uncertainty-aware confidence region. Clustering analysis is then performed on the uncertain set of instances, and the resulting prototypes are further integrated with those generated by LIFT. Despite the benefits of LIFT in enhancing instance representations in MLL/PLL tasks, two limitations arise when it is directly applied to LDL. First, the collected prototypes, i.e., cluster centers, primarily capture the intrinsic relationships of instances within individual clusters, while completely neglecting interactions across different clusters and structural correlations among prototypes. Second, the LSFs constructed by LIFT depend solely on a single distance metric, specifically the Euclidean distance, which may introduce noise and bias into resulting representations. As LDL involves a richer and more fine-grained label space than MLL or PLL, it calls for a more expressive and discriminative label-specific feature space. These limitations highlight the need for a more informative and structure-aware strategy for constructing LSFs in LDL tasks.

To address these issues, we argue that label-specific instance representations in LDL should move beyond isolated prototypes by explicitly modeling the structural dependencies and spatial relationships among them. Traditional prototype-only representations often ignore the relative geometric layout and inter-prototype correlations, which become particularly important when instances lie near the boundaries of positive, negative, and uncertain classes, or fall between multiple clusters across these categories. To better capture the intrinsic geometry of each class, one promising solution is to identify a set of representative points that approximate the manifold formed by its prototypes. These points help preserve intra-class structural information and support a smoother, more continuous representation space, thereby alleviating the limitations of discrete prototype views. Moreover, beyond instance-to-representative point distances, incorporating directional information can reveal whether an instance is spatially aligned with the internal class structure. This offers a complementary perspective, relying on a single scale, that helps resolve ambiguities when Euclidean distances are similar. For example, two instances may be equidistant from a representative point but lie in opposite directions; directional features can effectively distinguish such cases, thereby improving class separability.

In this paper, we introduce \emph{\textbf{S}tructural \textbf{A}nchor \textbf{P}oints} (SAPs) to capture interactions across different clusters and propose a novel LSFs construction strategy, \textbf{L}abel-specif\textbf{I}c \textbf{F}ea\textbf{T}ure with \textbf{SAP}s (LIFT-SAP). Specifically, for each label, we divide LDL training instances into positive, negative, and uncertain sets and gather prototypes by independently performing clustering analysis on these sets. Subsequently, SAPs are determined by computing the midpoints between each pair of prototypes within each set. LIFT-SAP re-characterizes each instance by not only querying its distances to all prototypes but also integrating its distance and directional information relative to SAPs. Furthermore, to the best of our knowledge, no existing LDL algorithm effectively utilizes multiple LSF spaces. Thus, we design a novel LDL algorithm, i.e., \textbf{L}abel \textbf{D}istribution \textbf{L}earning via \textbf{L}abel-specif\textbf{I}c \textbf{F}ea\textbf{T}ure with \textbf{SAP}s, short for LDL-LIFT-SAP, which employs a two-step learning strategy to unify multiple label description degrees predicted from different LSF spaces into a cohesive label distribution. Extensive experiments on $15$ LDL datasets demonstrate the effectiveness of LIFT-SAP over LIFT, as well as the superiority of LDL-LIFT-SAP compared to seven other well-established algorithms.

In summary, our contributions are as follows.
\begin{itemize}
    \item SAPs are introduced to capture interactions across different clusters, effectively exploring structural correlations among prototypes;
    \item LIFT-SAP is proposed to construct more comprehensive LSFs, integrating the distance and directional information of instances relative to SAPs;
    \item LDL-LIFT-SAP is designed to perform LDL from multiple LSF spaces, demonstrating its superiority across $15$ LDL tasks.
\end{itemize}

The remainder of this paper is organized as follows. Section \ref{sec:relatedwork} introduces related works on LDL and LSFs learning. In Sections \ref{sec:LIFT-SAP} and \ref{sec:LDL-LIFT-SAP}, the LIFT-SAP strategy and LDL-LIFT-SAP algorithm are respectively proposed in detail. Section \ref{sec:experiments} describes data sets, evaluation metrics, settings, and baselines, and then analyzes the results of comparative studies on $15$ real-world data sets. Finally, we conclude this paper in Section \ref{sec:conclusion}.

\section{Related Works}
\label{sec:relatedwork}
\subsection{Label Distribution Learning}
Numerous works have advanced LDL's performance and broadened its applications. Early integration with deep neural networks led to DLDL~\cite{gao-et-al:DLDL}, which utilizes deep features, while DLDLF~\cite{shen-et-al:LDLF} employs differentiable random forests to model label distributions. LSE-LDL~\cite{xu-et-al:latent-semantics-encoding} introduces latent semantics encoding to mitigate noise, and LDLSF~\cite{ren-et-al:LDLSF} jointly learns shared features across all labels alongside label-specific features for individual label. Other notable algorithms include MDLRML~\cite{tan-et-al:multilabel-distribution-learning}, which uses manifold modeling to capture the intrinsic topological structure of label distributions; CLDL~\cite{zhao-et-al:CLDL}, which examines the continuity of label distributions; and LDL-LRR~\cite{jia-et-al:LDL-LRR}, which leverages semantic relations among labels through label ranking information. Recent efforts have focused on capturing dependencies among different labels. EDL-LRL~\cite{jia-et-al:EDL-LRL} and LDL-SCL~\cite{jia-et-al:LDL-SCL} model local label correlations within clusters generated through clustering analysis in the label space, while LDLLC~\cite{jia-et-al:LDLLC} and LALOT~\cite{zhao-zhou:LALOT} leverage the Pearson correlation coefficient and ground metric learning methods, respectively, to capture global label correlations. Furthermore, LDL-LCLR~\cite{ren-et-al:LDL-LCLR} and LDL-LDM~\cite{wang-geng:LDL-LDM} simultaneously account for both local and global label correlations. Given the labor-intensive nature of complete LDL annotation, algorithms such as IncomLDL-LCD~\cite{xu-et-al:IncomLDL-LCD}, IncomLDL-prox~\cite{xu-et-al:IncomLDL}, and IncomLDL-admm~\cite{xu-et-al:IncomLDL} exploit these label correlations to address incomplete LDL tasks. The above LDL algorithms share a common characteristic of leveraging domain-specific insights, including continuity, ranking, sparsity or low-rank structures, and topology, to more effectively model real-world label distributions.

\subsection{Label-specific Features Learning}
LSFs aim to provide distinctive instance representations for each label. Generally, strategies for constructing LSFs can broadly be developed along two main lines, namely LSFs selection and LSFs transformation.

LSFs selection involves constructing LSFs by identifying the most pertinent and discriminative subset of features for each label from the original feature set. LLSF and LLSF-DL~\cite{huang-et-al:LLSF} leverage lasso regression to learn LSFs for MLL tasks while exploiting both second-order and high-order label correlations. CLLFS~\cite{xu-et-al:CLLFS} enhances the precision of LSFs selection by combating inaccuracies arising from false-positive labels, refining credible labels to enable accurate LSFs learning for each label and common feature learning across all labels. For PML tasks, LSNRLS~\cite{zou-et-al:LSNRLS} learns shared, non-redundant LSFs by integrating intrinsic feature correlations with learned label correlations, effectively mitigating the impact of noisy features.

LSFs transformation construct LSFs by mapping the original feature space into a unique and discriminative feature space tailored to each label. LIFT~\cite{zhang-wu:lift}, a pioneering work in prototype-based LSFs transformation for MLL tasks, gathers prototypes for each label by independently clustering positive and negative instances, then re-characterizes each instance based on its distances to all prototypes. CLIF~\cite{hang-et-al:CLIF} collaboratively learns deep LSFs and label semantics, while Deep-LIFT~\cite{li-et-al:Deep-LIFT} extracts deep LSFs by precisely aligning each label with its corresponding local visual region, achieving superior performance in image annotation tasks. WRAP \cite{Yu-et-al:WRAP} learns one (kernelized) linear model for each label, where LSFs are jointly constructed in an embedded feature space via empirical loss minimization and pairwise label correlation regularization. AME-LSIFT~\cite{Zhang-et-al:AME-LSIFT} employs c-means clustering to generate label subset-specific features, thereby enhancing the performance of individual base classifiers in multi-label ensemble learning. UCL~\cite{dong-et-al:label-specific} extends LIFT to PLL tasks by introducing an uncertainty-aware confidence region to accommodate false-positive labels. LDLSF~\cite{ren-et-al:LDLSF} simultaneously learns shared features across all labels and LSFs for each label to enhance the performance of LDL models. LDL-LDSF~\cite{wang-et-al:LDSFs} learns label-distribution-specific features by exploiting fuzzy cluster structures inherent in LDL tasks.

In summary, existing LDL works have achieved notable progress by leveraging deep architectures, modeling label correlations, and exploiting structural properties of label distributions. Meanwhile, LSF-based approaches, particularly prototype-based strategies such as LIFT, have proven effective in enhancing instance representations in MLL and PLL. However, the use of LSFs in LDL remains limited, and LIFT, in its existing form, overlooks structural correlations among prototypes and relies exclusively on Euclidean distance. These limitations motivate us to develop a structure-aware extension of LIFT that better captures the fine-grained label semantics in LDL tasks.

\section{The LIFT-SAP Strategy}
\label{sec:LIFT-SAP}

\subsection{Preliminaries}
Formally, let $\mathcal{X} = \mathbb{R}^{m}$ denote the $m$-dimensional input space and $\mathcal{Y} = \mathbb{R}^{p}$ denote the label space with $p$ labels $\{y_1,y_2,\ldots,y_p\}$. Given a set of $n$ training examples $\mathcal{D} = \{(\mathbf{x}_i,\mathcal{Y}_i)|1 \leq i \leq n\}$, where $\mathbf{x}_i \in \mathcal{X}$ is a $m$-dimensional feature vector, and $\mathcal{Y}_i \in \mathcal{Y}$ is a $p$-dimensional label distribution associated with $\mathbf{x}_i$ such that $\mathcal{Y}_i = [y_i^1,y_i^2,\ldots,y_i^p]$. For each label $y_j$ $(1 \leq j \leq p)$, $y_i^j \in [0,1]$ is the description degree of $y_j$ to $\mathbf{x}_i$, and $\sum\nolimits_{j=1}^p y_i^j=1$, which signifies that all $p$ labels can completely describe $\mathbf{x}_i$. 

The goal of LDL is to learn a mapping from input space $\mathcal{X}$ to label space $\mathcal{Y}$, i.e., $f: \mathcal{X} \rightarrow \mathcal{Y}$. Thus, for any unseen instance $\mathbf{x}'_i$, an LDL learner consider $f(\mathbf{x}'_i)$ as the predicted label distribution of $\mathbf{x}'_i$.

\subsection{LIFT for LDL}
LIFT aims to generate distinguishing features that effectively capture the specific characteristics of each $y_j \in \{y_1,y_2,\ldots,y_p\}$. To achieve this, LIFT leverages the intrinsic connections among instances across all $p$ labels. In MLL tasks, each $\mathbf{x}_i$ either owns $y_j$ or does not. Accordingly, Zhang et al.~\cite{zhang-wu:lift} categorize all training instances into positive and negative sets for each $y_j$. Extending LIFT to PLL tasks, Dong et al.~\cite{dong-et-al:label-specific} introduced a three-category division of training instances corresponding to each label, i.e., positive, negative, and uncertain sets, defined as follows:
\begin{align}
\mathcal{P}_j &= \{\mathbf{x}_i|(\mathbf{x}_i,\mathcal{Y}_i) \in \mathcal{D}, \tilde{y}_i^j > \tau_h\};\label{eq:1}\\
\mathcal{U}_j &= \{\mathbf{x}_i|(\mathbf{x}_i,\mathcal{Y}_i) \in \mathcal{D}, \tau_l \leq \tilde{y}_i^j \leq \tau_h \};\label{eq:2}\\
\mathcal{N}_j &= \{\mathbf{x}_i|(\mathbf{x}_i,\mathcal{Y}_i) \in \mathcal{D}, \tilde{y}_i^j < \tau_l\},
\label{eq:3}
\end{align}
where $\tilde{y}_i^j$ is the confidence level of $y_j$ being associated with $\mathbf{x}_i$, and $\tau_h$ and $\tau_l$ are predefined confidence threshold parameters.

Similarly, in LDL tasks, for each $y_j$, all training instances are divided into positive, negative, and uncertain sets, following the approach outlined in 
Eqs.~\eqref{eq:1}--\eqref{eq:3}. Notably, $\tilde{y}_i^j$ is essentially $y_i^j$, representing the degree to which $y_j$ describes $\mathbf{x}_i$. Consequently, $\tau_h$ and $\tau_l$ serve as threshold parameters for the description degree.

To gain insights into the properties of $\mathcal{P}_j$, $\mathcal{N}_j$, and $\mathcal{U}_j$, LIFT performs clustering analysis on each set. Following~\cite{dong-et-al:label-specific}, \emph{spectral clustering} is utilized to partition $\mathcal{P}_j$ into $m_j^+$ disjoint clusters with centers denoted as $C_{P^j} = \{\mathbf{cp}_1^j,\mathbf{cp}_2^j,...,\mathbf{cp}_{m_j^+}^j\}$, and $\mathcal{N}_j$ into $m_j^-$ disjoint clusters with centers denoted as $C_{N^j} = \{\mathbf{cn}_1^j,\mathbf{cn}_2^j,...,\mathbf{cn}_{m_j^-}^j\}$. Simultaneously, $\mathcal{U}_j$ is divided into $m_j^*$ disjoint clusters with centers denoted as $C_{U^j} = \{\mathbf{cu}_1^j,\mathbf{cu}_2^j,...,\mathbf{cu}_{m_j^*}^j\}$. Considering the prevalence of class-imbalance, LIFT equally treats the clustering information from both $\mathcal{P}_j$ and $\mathcal{N}_j$, while separately considering the number of clusters for $\mathcal{U}_j$. Specifically, the same number of clusters is assigned to $\mathcal{P}_j$ and $\mathcal{N}_j$ as follows:
\begin{equation}
m_j^+ = m_j^- = m_j = \lceil \sigma \cdot min(|\mathcal{P}_j|,|\mathcal{N}_j|)\rceil,
\label{eq:4}
\end{equation}
where $|\cdot|$ denotes the cardinality of a set, and $\sigma \in [0,1]$ is a ratio parameter controlling the number of clusters.

For $\mathcal{U}_j$, the number of clusters is set as follows:
\begin{equation}
m_j^* = \lceil \sigma \cdot |\mathcal{U}_j|\rceil.
\label{eq:5}
\end{equation}

Cluster centers serve to capture the underlying structures of $\mathcal{P}_j$, $\mathcal{N}_j$, and $\mathcal{U}_j$ with respect to $y_j$. Based on this, a mapping $\phi_j: \mathcal{X} \rightarrow LIFT_j$ is constructed, transforming the original $m$-dimensional input space $\mathcal{X}$ into a $(2m_j+m_j^*)$-dimensional LSF space as follows:
\begin{align}
\phi_j(\mathbf{x}_i) = 
\big[&d(\mathbf{x}_i,\mathbf{cp}_1^j),\ldots,d(\mathbf{x}_i,\mathbf{cp}_{m_j}^j),\nonumber\\
&d(\mathbf{x}_i,\mathbf{cn}_1^j),\ldots,d(\mathbf{x}_i,\mathbf{cn}_{m_j}^j),\nonumber\\
&\alpha \times d(\mathbf{x}_i,\mathbf{cu}_1^j),\ldots,\alpha \times d(\mathbf{x}_i,\mathbf{cu}_{m_j^*}^j) \big],
\label{eq:6}
\end{align}
where $d(\cdot,\cdot)$ returns the distance between two instances and is set to Euclidean metric. The parameter $\alpha$ is a discount factor to diminish the influence of uncertain instances.

The LSFs constructed by LIFT have proven effective in handling MLL~\cite{zhang-wu:lift,xu-et-al:LIFT-FRS} and PLL~\cite{dong-et-al:label-specific} tasks. Nevertheless, we observe two limitations that may impact LIFT's performance. Firstly, while cluster centers serve as foundational prototypes for the construction of LSFs, they primarily focus on the intrinsic relationships of instances within individual clusters, neglecting interactions across different clusters and structural correlations among prototypes. Secondly, as shown in Eq.~\eqref{eq:6}, the LSFs construction with LIFT relies solely on distance information, i.e., Euclidean metric. A multi-perspective approach could provide a more comprehensive characterization of instances, helping to mitigate noise and bias that may arise from depending on a single Euclidean metric.

\subsection{Structural Anchor Point}
To address the above limitations, we introduce SAPs to capture interactions across different clusters. Since the goal of constructing LSFs is to enhance the discrimination of the feature space w.r.t each label in learning tasks, we seek SAPs separately for different sets of instances. Specifically, for the positive set of instances $\mathcal{P}_j$, the SAPs, denoted as $SAP_{P^j} = \{\mathbf{sp}_1^j,\mathbf{sp}_2^j,...,\mathbf{sp}_{t_j}^j\}$, are defined as the midpoints between each pair of cluster centers in $\mathcal{P}_j$. Formally, we have:
\begin{equation}
SAP_{P^j} = \{\frac{\mathbf{cp}_{k_1}^j + \mathbf{cp}_{k_2}^j}{2}|\mathbf{cp}_{k_1}^j,\mathbf{cp}_{k_2}^j \in C_{P^j}\}.
\label{eq:7}
\end{equation}

Similarly, $SAP_{N^j} = \{\mathbf{sn}_1^j,\mathbf{sn}_2^j,...,\mathbf{sn}_{t_j}^j\}$ in $\mathcal{N}_j$ and $SAP_{U^j} = \{\mathbf{su}_1^j,\mathbf{su}_2^j,...,\mathbf{su}_{t_j^*}^j\}$ in $\mathcal{U}_j$ are respectively defined as:
\begin{align}
SAP_{N^j} &= \{\frac{\mathbf{cn}_{k_1}^j + \mathbf{cn}_{k_2}^j}{2}|\mathbf{cn}_{k_1}^j,\mathbf{cn}_{k_2}^j \in C_{N^j}\};\\
SAP_{U^j} &= \{\frac{\mathbf{cu}_{k_1}^j + \mathbf{cu}_{k_2}^j}{2}|\mathbf{cu}_{k_1}^j,\mathbf{cu}_{k_2}^j \in C_{U^j}\}.
\label{eq:9}
\end{align}

Furthermore, it's obvious that the number of SAPs in both $SAP_{P^j}$ and $SAP_{N^j}$ is $t_j = m_j \cdot (m_j-1)/2$, while the number of SAPs in $SAP_{U^j}$ is $t_j^* = {m_j^*} \cdot ({m_j^*}-1)/2$.

\subsection{LSFs with SAPs}
We construct LSFs by utilizing SAPs from two perspectives, i.e., distance and directional information, within the LIFT framework. For any instance $\mathbf{x}_i$, we first calculate the Euclidean distance between $\mathbf{x}_i$ and each SAP $\mathbf{sp}_k^j \in SAP_{P^j}$, $\mathbf{sn}_k^j \in SAP_{N^j}$, and $\mathbf{su}_{k'}^j \in SAP_{U^j}$, and a $(2t_j+t_j^*)$-dimensional LSF space is constructed via a mapping $\chi_j$, defined as follows:
\begin{align}
\chi_j(\mathbf{x}_i) = 
\big[&d(\mathbf{x}_i,\mathbf{sp}_1^j),\ldots,d(\mathbf{x}_i,\mathbf{sp}_{t_j}^j),\nonumber\\
&d(\mathbf{x}_i,\mathbf{sn}_1^j),\ldots,d(\mathbf{x}_i,\mathbf{sn}_{t_j}^j),\nonumber\\
&\alpha \times d(\mathbf{x}_i,\mathbf{su}_1^j),\ldots, \alpha \times d(\mathbf{x}_i,\mathbf{su}_{t_j^*}^j) \big],
\label{eq:10}
\end{align}
where $d(\mathbf{x}_i,\mathbf{sp}_k^j) = ||\mathbf{x}_i - \mathbf{sp}_k^j||$, $d(\mathbf{x}_i,\mathbf{sn}_k^j) = ||\mathbf{x}_i - \mathbf{sn}_k^j||$, $k = 1,2,...,t_j$, and $d(\mathbf{x}_i,\mathbf{su}_{k'}^j) = ||\mathbf{x}_i - \mathbf{su}_{k'}^j||$, ${k'} = 1,2,...,{t_j^*}$. The parameter $\alpha$ is a discount factor to diminish the influence of uncertain instances.

To complement the features derived from distance information and to mitigate potential noise and bias introduced by relying exclusively on Euclidean distances, we construct an additional $(2t_j+t_j^*)$-dimensional LSF space that focuses on directional information. Specifically, for any instance $\mathbf{x}_i$, we calculate the cosine of the angle between the vector $\mathbf{x}_i$ and each vector $\mathbf{sp}_k^j$, between $\mathbf{x}_i$ and each vector $\mathbf{sn}_k^j$, as well as between $\mathbf{x}_i$ and each vector $\mathbf{su}_{k'}^j$, to capture the directional connections of $\mathbf{x}_i$ with respect to each element in $SAP_{P^j}$, $SAP_{N^j}$, and $SAP_{U^j}$. Subsequently, a mapping $\psi_j$ is defined as follows:
\begin{align}
\psi_j(\mathbf{x}_i) = 
\big[&d_{cos}(\mathbf{x}_i,\mathbf{sp}_1^j),\ldots,d_{cos}(\mathbf{x}_i,\mathbf{sp}_{t_j}^j),\nonumber\\
&d_{cos}(\mathbf{x}_i,\mathbf{sn}_1^j),\ldots,d_{cos}(\mathbf{x}_i,\mathbf{sn}_{t_j}^j),\nonumber\\
&\alpha \times d_{cos}(\mathbf{x}_i,\mathbf{su}_1^j),\ldots,\alpha \times d_{cos}(\mathbf{x}_i,\mathbf{su}_{t_j^*}^j)\big],
\label{eq:11}
\end{align}
where
\begin{equation}
d_{cos}(\mathbf{x}_i,\mathbf{sp}_k^j) = \frac{\mathbf{x}_i \cdot \mathbf{sp}_k^j}{||\mathbf{x}_i|| \cdot ||\mathbf{sp}_k^j||},~ k = 1,2,...,t_j.
\label{eq:12}
\end{equation}
Moreover, $d_{cos}(\mathbf{x}_i,\mathbf{sn}_k^j)$ and $d_{cos}(\mathbf{x}_i,\mathbf{su}_{k'}^j)$ are defined similarly to $d_{cos}(\mathbf{x}_i,\mathbf{sp}_k^j)$, ${k'} = 1,2,...,{t_j^*}$.

\begin{figure*}[!t]
\centering
\subfloat[LIFT]{\includegraphics[width=1.5in]{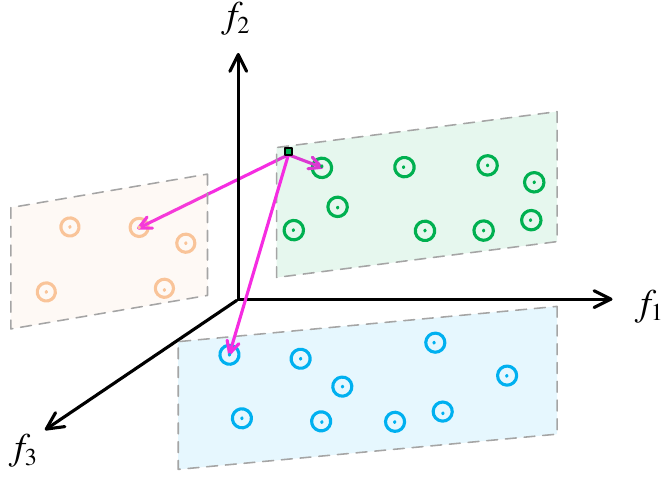}%
\label{fig:2a}}
\hfil
\subfloat[LIFT-SAP (distance)]{\includegraphics[width=1.5in]{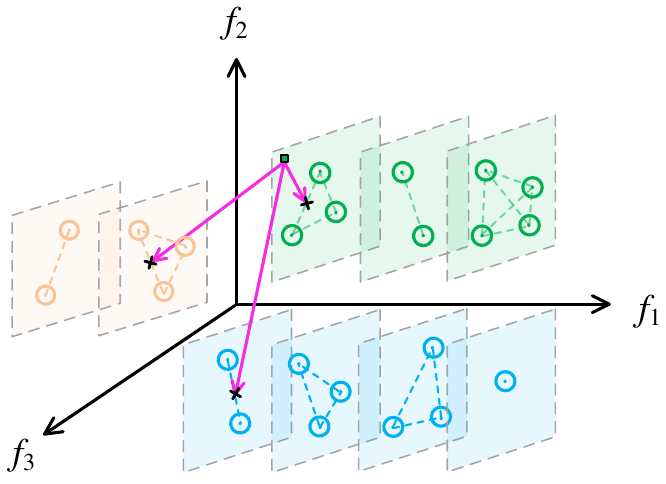}%
\label{fig:2b}}
\hfil
\subfloat[LIFT-SAP (direction)]{\includegraphics[width=1.5in]{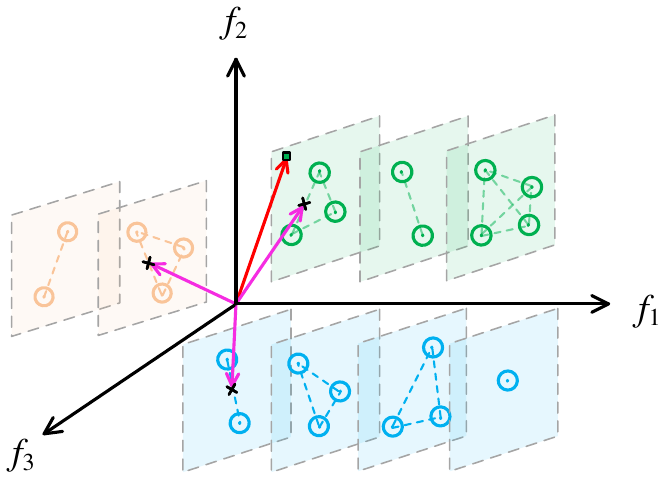}%
\label{fig:2c}}
\caption{Three types of LSFs constructed by (a) LIFT. (b) LIFT-SAP with distance information. (c) LIFT-SAP with directional information.}
\label{fig:2}
\end{figure*}

For better clarity, Fig.\ref{fig:2} illustrates three types of LSFs for a given label. The green, blue, and orange regions represent the positive, negative, and uncertain sets of training instances, with corresponding circles denoting clusters in these regions and dots marking their cluster centers (prototypes).

\begin{itemize}
  \item In LIFT, LSFs are constructed by calculating distances between a training instance and each cluster center. In Fig.3(a), the norm of each pink vector represents one of the LSFs constructed by LIFT. 
  \item In LIFT-SAP with distance information relative to SAPs, SAPs are identified by calculating midpoints between each pair of cluster centers within each set of training instances. LSFs are then constructed by calculating distances between a training instance and each SAP. In Fig.3(b), black crosses represent SAPs, and the norm of each pink vector represents one of the LSFs constructed by LIFT-SAP with distance information.
  \item In LIFT-SAP with directional information relative to SAPs, LSFs are constructed by calculating directions of a training instance to each SAP. In Fig.3(c), and the cosine of the angle between each pink vector and the red vector represents one of the LSFs constructed by LIFT-SAP with directional information.
\end{itemize}

\subsection{Fusing LSFs with SAPs into LIFT}
Our LIFT-SAP integrates the two newly constructed LSFs, i.e., $\chi_j(\mathbf{x}_i)$ and $\psi_j(\mathbf{x}_i)$, into the original LSFs $\phi_j(\mathbf{x}_i)$ within the LIFT framework. Here, we employ the simplest serial fusion means, and the LSFs, constructed via LIFT-SAP, are presented as follows:
\begin{equation}
{LIFT\_SAP}_j(\mathbf{x}_i) = [\lambda \phi_j(\mathbf{x}_i), \mu \chi_j(\mathbf{x}_i), \varepsilon \psi_j(\mathbf{x}_i)],
\label{eq:13}
\end{equation}
where $\lambda$, $\mu$, and $\varepsilon$ are balancing parameters that control the importance of the intrinsic relationships within individual clusters, the interactions across different clusters (capturing distance information), and the interactions across different clusters (capturing directional information), respectively.

For each label $y_j$, LIFT-SAP transforms the original $m$-dimensional input space $\mathcal{X}$ to a $(2{m_j}^2 + {m_j^*}^2)$-dimensional LSF space ${LIFT\_SAP}_j$. However, we noticed that the dimensionality of ${LIFT\_SAP}_j$ grows exponentially compared to that of ${LIFT}_j$, leading to a redundancy of information and a significant rise in computational overhead for subsequent classification learning. Therefore, in practice, since interactions across different clusters typically occur within a local area, we further divide the clusters into several disjoint blocks in $\mathcal{P}_j$, $\mathcal{N}_j$, and $\mathcal{U}_j$, respectively. And then, we consider only the midpoints between each pair of cluster centers within the same block as SAPs. These disjoint blocks can be generated through a secondary \emph{$k$-means clustering} of the cluster centers in $\mathcal{P}_j$, $\mathcal{N}_j$, and $\mathcal{U}_j$, respectively. Ideally, $k$ is set to $\lceil\frac{m_j}{4}\rceil$ or $\lceil\frac{m_j}{5}\rceil$ for $\mathcal{P}_j$ and $\mathcal{N}_j$, and to $\lceil\frac{m_j^*}{4}\rceil$ or $\lceil\frac{m_j^*}{5}\rceil$ for $\mathcal{U}_j$, to ensure that the average number of cluster centers per block is approximately $4$ or $5$.

Take Fig.\ref{fig:2} as an example once again. In Fig.3(a), the clusters within the green, blue, and orange regions are divided into three, four, and two disjoint blocks, respectively, in LIFT-SAP, as illustrated in Fig.3(b) and Fig.3(c). SAPs are considered only between cluster centers within the same block, enabling a localized search that minimizes information redundancy and significantly enhances computational efficiency.

\begin{algorithm}[tb]
    \caption{Label-specifIc FeaTure with SAPs (LIFT-SAP)}
    \label{alg:algorithm1}
    \textbf{Input}:\\
    $\mathcal{D}$: LDL training examples $\{(\mathbf{x}_i,\mathcal{Y}_i)|1 \leq i \leq n\}$ \\
    ($\mathbf{x}_i \in \mathcal{X}$, $\mathcal{Y}_i \in \mathcal{Y}$ with $p$ labels $\{y_1,y_2,\ldots,y_p\}$)\\
    \textbf{Parameter}:\\
    $\tau_h$, $\tau_l$: confidence thresholds as used in Eqs.~\eqref{eq:1}--\eqref{eq:3}\\
    $\sigma$: ratio parameter as used in Eq.~\eqref{eq:4} and Eq.~\eqref{eq:5}\\
    $\alpha$: discount factor as used in Eq.~\eqref{eq:6}, Eq.~\eqref{eq:10}, and Eq.~\eqref{eq:11}\\
    $\lambda$, $\mu$, $\varepsilon$: balancing parameters as used in Eq.~\eqref{eq:13}\\
    \textbf{Output}: $\{T_1^*, T_2^*,\dots,T_p^*\}$
    \begin{algorithmic}[1] 
        \STATE \textbf{for} $j = 1$ to $p$ \textbf{do}
        \STATE \quad Form $\mathcal{P}_j$, $\mathcal{U}_j$, and $\mathcal{N}_j$ based on $\mathcal{D}$ by Eqs.~\eqref{eq:1}--\eqref{eq:3};
        \STATE \quad Perform \emph{spectral clustering} on $\mathcal{P}_j$ and $\mathcal{N}_j$, each with\\
               \quad $m_j$ clusters as Eq.~\eqref{eq:4}, and on $\mathcal{U}_j$ with $m_j^*$ clusters as\\
               \quad Eq.~\eqref{eq:5};
        \STATE \quad Create the mapping $\phi_j$ for $y_j$ by Eq.~\eqref{eq:6};
        \STATE \quad Seek SAPs on $\mathcal{P}_j$, $\mathcal{N}_j$ and $\mathcal{U}_j$ by Eqs.~\eqref{eq:7}--\eqref{eq:9};
        \STATE \quad Create the mapping $\chi_j$ for $y_j$ by Eq.~\eqref{eq:10};
        \STATE \quad Create the mapping $\psi_j$ for $y_j$ by Eq.~\eqref{eq:11};
        \STATE \quad Construct ${LIFT\_SAP}_j$ via fusing $\chi_j$ and $\psi_j$ into $\phi_j$\\
               \quad by Eq.~\eqref{eq:13};
        \STATE \quad Form $T_j^*$ by Eq.~\eqref{eq:14};
        \STATE \textbf{end for}
        \STATE \textbf{return} $\{T_1^*, T_2^*,\dots,T_p^*\}$
    \end{algorithmic}
\end{algorithm}

\begin{remark}
It is worth noting that we do not claim that the midpoints between each pair of cluster centers are the best possible realization for identifying the SAPs in LIFT-SAP. Indeed, exploring certain cluster properties, such as \textit{class imbalance among clusters} and \textit{differences in cluster density}, might further benefit to capturing interactions across different clusters and establishing structural correlations among prototypes. Nevertheless, the simplest and most direct way adopted in LIFT-SAP suffices to yield competitive performances as shown in Sec.\ref{sec:experiments}.
\end{remark}

Algorithm \ref{alg:algorithm1} outlines the pseudo-code of LIFT-SAP strategy. Its computational complexity can be decomposed into three main components. For each label $y_j$, spectral clustering over the positive, negative, and uncertain subsets has a complexity of $\mathcal{O}\big(n^2(2m_j+m_j^*)\big)$. Constructing the mapping $\phi_j$ requires $\mathcal{O}\big(n(2m_j+m_j^*)m\big)$ operations, while constructing the mappings $\chi_j$ and $\psi_j$ requires $\mathcal{O}\Big(n\big(2m_j(m_j-1) + m_j^*(m_j^*-1)\big)m\Big)$. Combining these, the overall complexity across all labels is: $\mathcal{O}\Big(\sum\limits_{j=1}^p \big(n^2(2m_j+m_j^*) + n(2m_j^2+{m_j^*}^2)m\big)\Big)$, where, for each label $y_j$, $m_j$ denotes the number of clusters in its corresponding positive or negative subset, and $m_j^*$ denotes the number of clusters in the uncertain subset.

\section{The LDL-LIFT-SAP Algorithm}
\label{sec:LDL-LIFT-SAP}
LIFT-SAP independently constructs a family of $p$ LSF spaces $\{{LIFT\_SAP}_1,{LIFT\_SAP}_2,\ldots,$ ${LIFT\_SAP}_p\}$ to enhance feature discrimination for each label. Formally, for each $y_j$, a training set $T_j^*$ comprising $n$ instances is derived from the set of training examples $\mathcal{D}$ as follows:
\begin{equation}
T_j^* = \{({LIFT\_SAP}_j(\mathbf{x}_i),{y}_i^j)| (\mathbf{x}_i,\mathcal{Y}_i) \in \mathcal{D}\}.
\label{eq:14}
\end{equation}

\begin{figure*}[!t]
\centering
\includegraphics[width=0.85\textwidth]{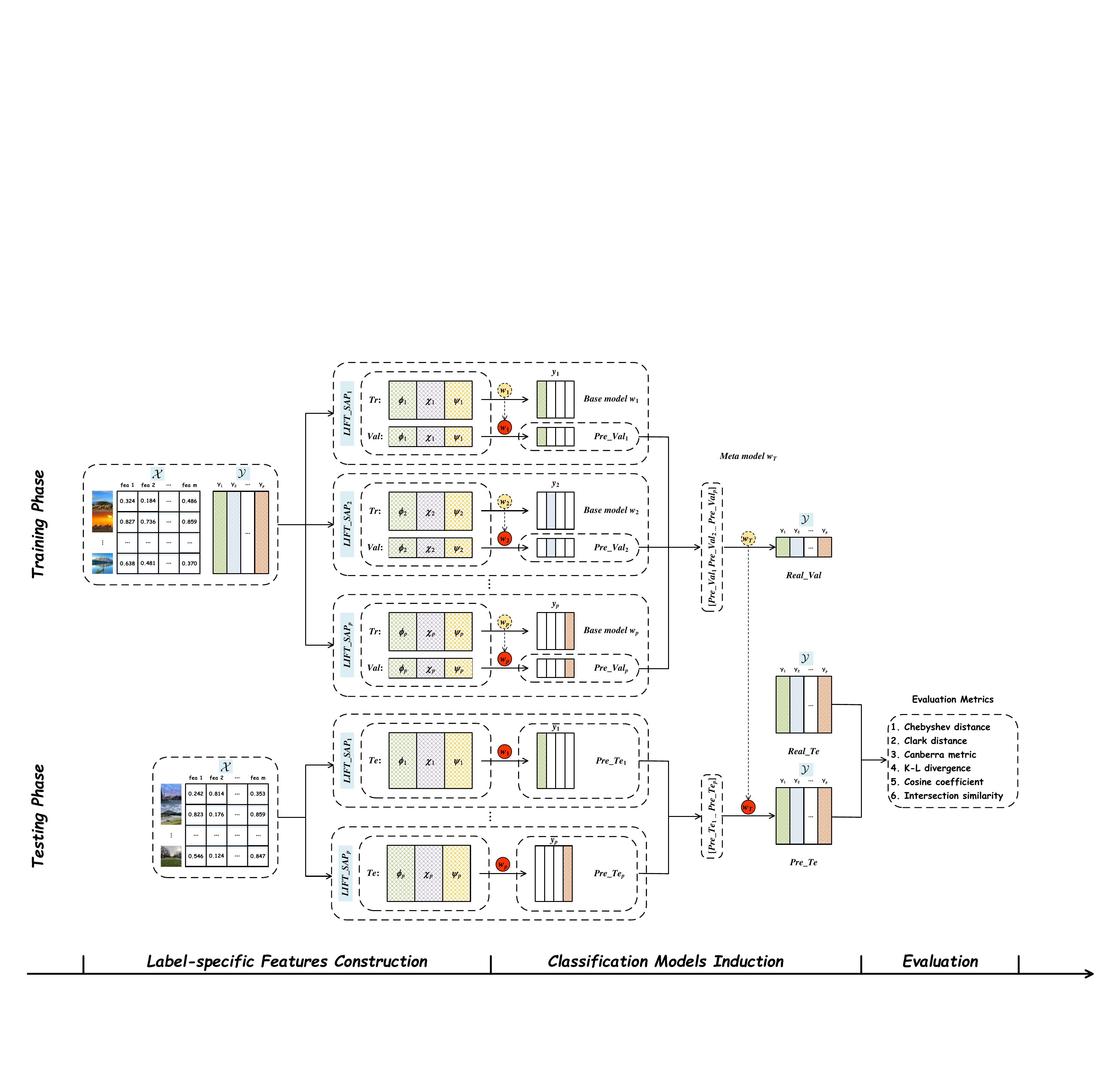}
\caption{Label Distribution Learning via Label-specifIc FeaTure with Structural Anchor Points (LDL-LIFT-SAP)}
\label{fig:3}
\end{figure*}

Unlike in MLL, where LIFT treats different labels independently as each instance can be associated with multiple ground-truth labels, predicting each label separately within its respective LIFT-SAP-constructed LSF space is unfeasible for LDL tasks. This limitation arises from the necessity for LDL classification models to adhere to inter-label constraints, ensuring that all labels collectively and completely describe each instance. Hence, we present a novel algorithm, \textbf{L}abel \textbf{D}istribution \textbf{L}earning via \textbf{L}abel-specif\textbf{I}c \textbf{F}ea\textbf{T}ure with \textbf{SAP}s (LDL-LIFT-SAP), capable of integrating label description degrees predicted from multiple LSF spaces. Fig.\ref{fig:3} is the overall framework of LDL-LIFT-SAP.

\textbf{Training phase:} For each label $y_j$ $(1 \leq j \leq p)$, a LSF space ${LIFT\_SAP}_j$ is constructed, consisting of three types of LSFs: $\phi_j$, $\chi_j$, and $\psi_j$. Each ${LIFT\_SAP}_j$ encapsulates distinctive information specific to label $y_j$, enabling more effective modeling of the dependencies between $\mathcal{X}$ and $y_j$. The training instances are then further divided into a training group ($Tr$) and a validation group ($Val$). A base model $w_j$ is trained on $Tr$ within each ${LIFT\_SAP}_j$ to predict the description degrees of label $y_j$, outputting the predictions $Pre\_Val_j$ for $Val$. Repeating this process for all $p$ labels yields a collection of prediction outputs $\{Pre\_Val_1,Pre\_Val_2,...,Pre\_Val_p\}$ for $Val$. These outputs, treated as the second-level features of $Val$, are combined with the real description degrees of all $p$ labels, denoted as $Real\_Val$, to train a meta model $w_T$ for predicting the description degrees of all $p$ labels.

\textbf{Testing phase:} For unseen instances in a testing group ($Te$), LSF representations ${LIFT\_SAP}_j(\mathbf{x}_u)$, where $\mathbf{x}_u \in \mathcal{X}$ in $Te$ and $1 \leq j \leq p$, are constructed using the same way as in the training phase. The trained base model $w_j$ then predicts the description degrees of label $y_j$ for all $\mathbf{x}_u \in \mathcal{X}$ in $Te$, yielding the outputs $Pre\_Te_j$. Repeating this for all $p$ labels results in a collection of prediction outputs $\{Pre\_Te_1,Pre\_Te_2,...,Pre\_Te_p\}$ for $Te$, which is treated as the second-level features of $Te$. Finally, the trained meta model $w_T$ predicts the final description degrees of all $p$ labels, denoted as $Pre\_Te$.

\begin{algorithm}[tb]
    \caption{Label Distribution Learning via Label-specifIc FeaTure with SAPs (LDL-LIFT-SAP)}
    \label{alg:algorithm2}
    \textbf{Input}:\\
    $T_j^*$: LDL training sets transformed by LIFT-SAP \\
    $\{({LIFT\_SAP}_j(\mathbf{x}_i),{y}_i^j)|1 \leq i \leq n, 1 \leq j \leq p\}$\\
    $\mathbf{x}_u$: unseen instance $\mathbf{x}_u \in \mathcal{X}$\\
    \textbf{Output}: label distribution $Pre(\mathbf{x}_u)$ for $\mathbf{x}_u$\\
    \vspace{-5mm}
    \begin{algorithmic}[1] 
        \STATE \textbf{for} $j = 1$ to $p$ \textbf{do}
        \STATE \quad Divide $T_j^*$ into the training group ($Tr$) and the \\
               \quad validation group ($Val$) as:\\
               \quad $Tr$: $\{({LIFT\_SAP}_j(\mathbf{x}_i),{y}_i^j)|1 \leq i \leq n_{Tr}\}$,\\
               \quad $Val$: $\{({LIFT\_SAP}_j(\mathbf{x}_i),{y}_i^j)|n_{Tr} \leq i \leq n\}$;\\
        \STATE \quad Train a base model $w_j$ on $Tr$;
        \STATE \quad Predict $Pre\_Val_j$ for $Val$ by $w_j$;
        \STATE \textbf{end for}
        \STATE Train a meta model $w_T$ on $[Pre\_Val_1,...,Pre\_Val_p]$
               and $[{y}_i^1,{y}_i^2,...,{y}_i^p]$, where $n_{Tr} \leq i \leq n$;
        \STATE \textbf{for} $j = 1$ to $p$ \textbf{do}
        \STATE \quad Transform $\mathbf{x}_u$ into ${LIFT\_SAP}_j(\mathbf{x}_u)$ by LIFT-SAP;
        \STATE \quad Predict $Pre(\mathbf{x}_u)_j$ for $\mathbf{x}_u$ by $w_j$;
        \STATE \textbf{end for}
        \STATE Predict $Pre(\mathbf{x}_u)$ for $[Pre(\mathbf{x}_u)_1,...,Pre(\mathbf{x}_u)_p]$ by $w_T$;
        \STATE \textbf{return} label distribution $Pre(\mathbf{x}_u)$ for $\mathbf{x}_u$
    \end{algorithmic}
\end{algorithm}

Algorithm \ref{alg:algorithm2} outlines the pseudo-code of LDL-LIFT-SAP algorithm. Its computational complexity can be decomposed into three main components. For each label $y_j$, the training of its corresponding base model has a computational cost of $\mathcal{O}(C_{base}^j)$, while the construction of second-level features requires $\mathcal{O}\big(n(2m_j^2+{m_j^*}^2)\big)$ operations. The training of the meta model has a computational cost of $\mathcal{O}(C_{meta})$. Therefore, the overall complexity is: $\mathcal{O}\Big(\sum\limits_{j=1}^p \big(C_{base}^j + n(2m_j^2+{m_j^*}^2)\big) + C_{meta}\Big)$, where, for each label $y_j$, $m_j$ denotes the number of clusters in its corresponding positive or negative subset, and $m_j^*$ denotes the number of clusters in the uncertain subset.

\section{Experiments}
\label{sec:experiments}

\subsection{Experimental Configuration}
\subsubsection{Datasets} 
We employ $15$ publicly available datasets from the LDL repository\footnote{\url{https://palm.seu.edu.cn/xgeng/LDL/#data}}. Tab.~\ref{tab:1} summarizes some statistics of these datasets, with $\#Inst.$, $\#Feat.$, and $\#Lab.$ denoting the number of instances, features, and labels, respectively. The datasets were chosen from four distinct practical application domains, i.e., bioinformatics, natural scene recognition, facial expression recognition, and facial beauty assessment. 

The first nine datasets originate from yeast gene biological experiments~\cite{EisenM}, each containing $2,465$ genes characterized by $24$-dimensional phylogenetic profile vectors. For these datasets, the label distributions are derived from normalized gene expression levels measured at discrete time points during biological experiments.

The \emph{Natural Scene} dataset ($10$th) comprises $2,000$ images, each represented by a $294$-dimensional feature vector extracted using Boutell et al.'s method~\cite{BoutellM}. Nine scene labels (plant, sky, cloud, snow, building, desert, mountain, water, and sun) were independently ranked by ten human evaluators based on relevance. These rankings were subsequently transformed into label distributions through nonlinear programming, optimized using the log barrier interior-point method.

The \emph{S-JAFFE} dataset ($11$th) contains $213$ grayscale facial expression images from $10$ Japanese female models, with features extracted using Local Binary Patterns (LBP)~\cite{AhonenT} yielding $243$-dimensional vectors. The \emph{S-BU\_3DFE} dataset ($12$th) includes $2,500$ facial expression images. Both datasets utilize Ekman's six basic emotions (anger, disgust, fear, joy, sadness, and surprise), with emotion intensities assessed by $60$ and $23$ evaluators respectively. The normalized average scores constitute the label distributions.

For facial beauty assessment, the \emph{M$^2$B} dataset ($13$th) contains $1,240$ facial images ($128 \times 128$ pixels), with features extracted using a combination of LBP, Gabor Filter (GF)~\cite{JainAK}, and Color Moment (CM), dimensionally reduced to $250$ through PCA. Beauty distributions were generated using a $k$-wise comparison strategy ($k=10$), with each image evaluated by at least $15$ assessors across five beauty levels~\cite{NguyenTV}. The \emph{SCUT-FBP} dataset ($14$th) comprises $1,500$ facial images ($350 \times 350$ pixels), featuring LBP, Histogram of Oriented Gradient (HOG), and GF descriptors reduced to $300$ dimensions~\cite{XieD}. Beauty distributions were derived from $75$ assessors' ratings across five attractiveness levels.

The \emph{Emotions6} dataset ($15$th) contains $1,980$ facial expression images collected from Flickr~\cite{PengK}. Features were extracted using LBP, HOG, and CM, with dimensionality reduced to $168$ through PCA. This dataset extends beyond Ekman's six basic emotions to include a `neutral' state, with emotional responses collected from $15$ subjects via Amazon Mechanical Turk.

\begin{table}[!t]
\setlength{\tabcolsep}{3.5pt}
\caption{Characteristics of the $15$ experimental datasets.}
\centering
\begin{tabular}{ccccccccccc}
    \hline
    ID & Dataset & \#Inst. & \#Feat. & \#Lab. & Domain \\
    \hline
    $1$ & Yeast-alpha & $2$,$465$ & $24$ & $18$ & Bioinformatics\\
    $2$ & Yeast-cdc & $2$,$465$ & $24$ & $15$ & Bioinformatics\\
    $3$ & Yeast-cold & $2$,$465$ & $24$ & $4$ & Bioinformatics\\
    $4$ & Yeast-diau & $2$,$465$ & $24$ & $7$ & Bioinformatics\\
    $5$ & Yeast-dtt & $2$,$465$ & $24$ & $4$ & Bioinformatics\\
    $6$ & Yeast-heat & $2$,$465$ & $24$ & $6$ & Bioinformatics\\
    $7$ & Yeast-spo & $2$,$465$ & $24$ & $6$ & Bioinformatics\\
    $8$ & Yeast-spo5 & $2$,$465$ & $24$ & $3$ & Bioinformatics\\
    $9$ & Yeast-spoem & $2$,$465$ & $24$ & $2$ & Bioinformatics\\
    $10$ & Natural Scene & $2$,$000$ & $294$ & $9$ & Natural scene recognition\\
    $11$ & S-JAFFE & $213$ & $243$ & $6$ & Facial expression recognition\\
    $12$ & S-BU\_3DFE & $2$,$500$ & $243$ & $6$ & Facial expression recognition\\
    $13$ & M$^2$B & $1$,$240$ & $250$ & $5$ & Facial beauty assessment\\
    $14$ & SCUT-FBP & $1$,$500$ & $300$ & $5$ & Facial beauty assessment\\
    $15$ & Emotions6 & $1$,$980$ & $168$ & $7$ & Facial expression recognition\\
    \hline
\end{tabular}
\label{tab:1}
\end{table}

\subsubsection{Evaluation Metrics} 
Consistent with prior works~\cite{geng:LDL,xu-et-al:IncomLDL-LCD}, we employ six evaluation metrics to assess LDL algorithms, i.e., four distance-based metrics (\emph{lower values imply better performance, denoted as $\downarrow$}): \emph{Chebyshev distance}, \emph{Clark distance}, \emph{Canberra metric}, and \emph{Kullback-Leibler (K-L) divergence}; and two similarity-based metrics (\emph{higher values imply better performance, denoted as $\uparrow$}): \emph{Cosine coefficient} and \emph{Intersection similarity}. By capturing diverse aspects of label distribution differences, these metrics provide a comprehensive assessment for LDL algorithms.

For an unseen instance $\mathbf{x}_u \in \mathcal{X}$, the ground-truth label distribution is denoted as $\mathcal{Y}_u = [y_u^1,y_u^2,\ldots,y_u^p]$, while the label distribution predicted by an LDL algorithm is represented as $\overline{\mathcal{Y}_u} = [\overline{y_u^1},\overline{y_u^2},\ldots,\overline{y_u^p}]$, then six evaluation metrics are presented in Tab.~\ref{tab:2}.

\begin{table}[!t]
\caption{Evaluation metrics for label distribution learning}
\centering
\begin{tabular}{cc}
\hline
Metrics & Formulas \\
\hline
Chebyshev distance $\boldsymbol{\downarrow}$ & $\max\limits_{j} |\overline{y_u^j} - y_u^j|$ \\
Clark distance $\boldsymbol{\downarrow}$ & $\left( \sum\limits_{j=1}^{p} \frac{(\overline{y_u^j} - y_u^j)^2}{(\overline{y_u^j} + y_u^j)^2} \right)^{\frac{1}{2}}$ \\
Canberra metric $\boldsymbol{\downarrow}$ & $\sum\limits_{j=1}^{p} \frac{|\overline{y_u^j} - y_u^j|}{\overline{y_u^j} + y_u^j}$ \\
Kullback-Leibler divergence $\boldsymbol{\downarrow}$ & $\sum\limits_{j=1}^{p} y_u^j \ln \left( \frac{y_u^j}{\overline{y_u^j}} \right)$ \\
Cosine coefficient $\boldsymbol{\uparrow}$ & $\frac{\sum\limits_{j=1}^{p} \overline{y_u^j} y_u^j}{\left( \sum\limits_{j=1}^{p} (\overline{y_u^j})^2 \right)^{\frac{1}{2}} \left( \sum\limits_{j=1}^{p} (y_u^j)^2 \right)^{\frac{1}{2}}}$ \\
Intersection similarity $\boldsymbol{\uparrow}$ & $\sum\limits_{j=1}^{p} \min(\overline{y_u^j}, y_u^j)$ \\
\hline
\end{tabular}
\label{tab:2}
\end{table}

\subsubsection{Settings and Baselines}
LDL-LIFT-SAP is compared with seven state-of-the-art LDL algorithms: LALOT~\cite{zhao-zhou:LALOT}, LDLLC~\cite{jia-et-al:LDLLC}, EDL-LRL~\cite{jia-et-al:EDL-LRL}, LDLSF~\cite{ren-et-al:LDLSF}, LDL-LCLR~\cite{ren-et-al:LDL-LCLR}, LDL-SCL~\cite{jia-et-al:LDL-SCL}, and LDL-LDM~\cite{wang-geng:LDL-LDM}. These algorithm parameters are tuned according to their original references.

\begin{itemize}
  \item LALOT~\cite{zhao-zhou:LALOT}: Label correlation exploration is approached as ground metric learning with kernel-biased regularization, employing the optimal transport distance to quantify differences between predicted and real label distributions. The trade-off parameter $C$ and the entropic regularization coefficient $\lambda$ are determined through cross-validation.
  \item LDLLC~\cite{jia-et-al:LDLLC}: Global label correlation is modeled by calculating the distance between the corresponding columns of the parameter matrix. The trade-off parameters $\lambda_1$ and $\lambda_2$ are set to $10^{-1}$ and $10^{-2}$, respectively, with a maximum of $100$ iterations.
  \item EDL-LRL~\cite{jia-et-al:EDL-LRL}: Local label correlation is exploited by capturing the low-rank structure of clusters in label space using trace norm regularization. The trade-off parameters $\lambda_1$ and $\lambda_2$ are set to $10^{-3}$ and $10^{-2}$, respectively, with the number of clusters $m$ set to $5$.
  \item LDLSF~\cite{ren-et-al:LDLSF}: The common features for all labels and label-specific features for each label are simultaneously learned to enhance the LDL model. The trade-off parameters $\lambda_1$, $\lambda_2$, and $\lambda_3$ are set to $10^{-4}$, $10^{-2}$, and $10^{-3}$, respectively, and the penalty factor $\rho$ is set to $10^{-3}$.
  \item LDL-LCLR~\cite{ren-et-al:LDL-LCLR}: Global and local label correlations are jointly exploited by assuming that label correlations are globally low-rank and are exclusively shared and updated within local instances. The trade-off parameters $\lambda_1$, $\lambda_2$, $\lambda_3$, and $\lambda_4$ are set to $10^{-4}$, $10^{-3}$, $10^{-3}$, and $10^{-3}$, respectively, with the number of clusters $k$ set to $4$.
  \item LDL-SCL~\cite{jia-et-al:LDL-SCL}: Local label correlation is encoded as additional features based on clusters in the label space. Optimization is performed using Adaptive Moment Estimation (ADAM). The trade-off parameters $\lambda_1$, $\lambda_2$, and $\lambda_3$ are chosen from $10^{\{-3,-2,\dots,2,3\}}$, and the number of clusters is tuned from $0$ to $14$.
  \item LDL-LDM~\cite{wang-geng:LDL-LDM}: Global and local label correlations are simultaneously exploited by learning two types of label distribution manifolds. The trade-off parameter $\lambda_1$ is set to $10^{-3}$, while $\lambda_2$ and $\lambda_3$ are chosen from $10^{\{-4,-3,-2\}}$. The number of clusters is tuned from $1$ to $14$. 
\end{itemize}

For LIFT-SAP, the parameters $\lambda$, $\mu$, and $\varepsilon$ are tuned via grid search with values in $\{0, 0.05, \dots, 1\}$, subject to $\lambda + \mu + \varepsilon = 1$. The ratio parameter $\sigma$ and discount factor $\alpha$ are set to $0.1$ and $0.5$, respectively. Instead of directly setting $\tau_h$ and $\tau_l$, for each label, training instances with top $55\%$ description degrees are assigned to the positive set, the bottom $35\%$ to the negative set, and the rest to the uncertain set. Moreover, during the training phase of LDL-LIFT-SAP, we allocate $50\%$ of the training instances to the training group $Tr$, while the remaining instances are assigned to the validation group $Val$. The base models and the meta model are trained using the SA-BFGS~\cite{geng:LDL}.

\subsection{Comparative Studies}
For each dataset, $50\%$ of the examples are randomly sampled without replacement to form the training set, while the remaining $50\%$ are used as the testing set. This sampling process is repeated ten times, and the average predictive performance and the standard deviation across training/testing trials are recorded. The best performance is highlighted in bold.

\begin{table*}[!t]
\caption{Predictive performances on the $15$ LDL datasets evaluated by \texttt{Chebyshev distance} $\downarrow$.}
\centering
\scriptsize
\setlength{\tabcolsep}{6pt}
\begin{tabular}{ccccccccccc}
    \hline
    ID & LALOT & LDLLC & EDL-LRL & LDLSF & LDL-LCLR & LDL-SCL & LDL-LDM & LDL-LIFT-SAP\\
    \hline
    1  & 0.0138$\pm$0.0002 & 0.0136$\pm$0.0002 & 0.0136$\pm$0.0002 & 0.0135$\pm$0.0001 & 0.0135$\pm$0.0001 & 0.0135$\pm$0.0002 & \textbf{0.0134$\pm$0.0002} & \textbf{0.0134$\pm$0.0001} \\
    2  & 0.0168$\pm$0.0001 & 0.0168$\pm$0.0002 & 0.0168$\pm$0.0001 & \textbf{0.0162$\pm$0.0002} & 0.0163$\pm$0.0002 & 0.0163$\pm$0.0001 & \textbf{0.0162$\pm$0.0001} & \textbf{0.0162$\pm$0.0002} \\
    3  & 0.0551$\pm$0.0006 & 0.0538$\pm$0.0006 & 0.0535$\pm$0.0005 & 0.0514$\pm$0.0003 & 0.0514$\pm$0.0006 & \textbf{0.0509$\pm$0.0004} & 0.0514$\pm$0.0004 & 0.0513$\pm$0.0007 \\
    4  & 0.0409$\pm$0.0004 & 0.0410$\pm$0.0003 & 0.0412$\pm$0.0003 & 0.0372$\pm$0.0004 & 0.0374$\pm$0.0003 & \textbf{0.0368$\pm$0.0003} & 0.0371$\pm$0.0004 & \textbf{0.0368$\pm$0.0003} \\
    5  & 0.0381$\pm$0.0005 & 0.0369$\pm$0.0004 & 0.0368$\pm$0.0004 & 0.0362$\pm$0.0003 & \textbf{0.0361$\pm$0.0003} & \textbf{0.0361$\pm$0.0004} & 0.0362$\pm$0.0005 & \textbf{0.0361$\pm$0.0002} \\
    6  & 0.0440$\pm$0.0007 & 0.0425$\pm$0.0004 & 0.0427$\pm$0.0004 & 0.0426$\pm$0.0006 & 0.0424$\pm$0.0002  & 0.0424$\pm$0.0004 & \textbf{0.0423$\pm$0.0003} & \textbf{0.0423$\pm$0.0005} \\
    7  & 0.0615$\pm$0.0009 & 0.0605$\pm$0.0009 & 0.0601$\pm$0.0004 & 0.0589$\pm$0.0006 & 0.0586$\pm$0.0006 & \textbf{0.0577$\pm$0.0007} & 0.0588$\pm$0.0006 & 0.0586$\pm$0.0005 \\
    8  & 0.0922$\pm$0.0008 & 0.0924$\pm$0.0011 & 0.0913$\pm$0.0014 & 0.0919$\pm$0.0010 & 0.0916$\pm$0.0014 & \textbf{0.0908$\pm$0.0010} & 0.0915$\pm$0.0010 & 0.0918$\pm$0.0008 \\
    9 & 0.0887$\pm$0.0019 & 0.0875$\pm$0.0009 & 0.0874$\pm$0.0011 & 0.0874$\pm$0.0012 & 0.0872$\pm$0.0018 & 0.0871$\pm$0.0017 & \textbf{0.0868$\pm$0.0016} & 0.0874$\pm$0.0012 \\
    10 & 0.3674$\pm$0.0060 & 0.3689$\pm$0.0055 & 0.3889$\pm$0.0087 & 0.3316$\pm$0.0041 & 0.3862$\pm$0.0112 & 0.3292$\pm$0.0039 & 0.3325$\pm$0.0103 & \textbf{0.2750$\pm$0.0059} \\
    11 & 0.1189$\pm$0.0038 & 0.1306$\pm$0.0092 & 0.4443$\pm$0.0260 & 0.1624$\pm$0.0304 & 0.1044$\pm$0.0047 & 0.0952$\pm$0.0000 & 0.1074$\pm$0.0046 & \textbf{0.0915$\pm$0.0056} \\
    12 & 0.1381$\pm$0.0015 & 0.1084$\pm$0.0010 & 0.1169$\pm$0.0043 & 0.1127$\pm$0.0008 & 0.1105$\pm$0.0017 & 0.1178$\pm$0.0011 & 0.1080$\pm$0.0015 & \textbf{0.1025$\pm$0.0008} \\
    13 & 0.4977$\pm$0.0288 & 0.5429$\pm$0.0106 & 0.5982$\pm$0.0090 & 0.4124$\pm$0.0103 & 0.4351$\pm$0.0087 & 0.3793$\pm$0.0025 & 0.4267$\pm$0.0116 & \textbf{0.3584$\pm$0.0064} \\
    14 & 0.3755$\pm$0.0082 & 0.3876$\pm$0.0029 & 0.4227$\pm$0.0060 & 0.2975$\pm$0.0035 & 0.2624$\pm$0.0106 & 0.3036$\pm$0.0024 & 0.2597$\pm$0.0043 & \textbf{0.2433$\pm$0.0037} \\
    15 & 0.3454$\pm$0.0053 & 0.3306$\pm$0.0030 & 0.3510$\pm$0.0048 & \textbf{0.3101$\pm$0.0036} & 0.3177$\pm$0.0040 & 0.3127$\pm$0.0013 & 0.3121$\pm$0.0046 & \textbf{0.3101$\pm$0.0028} \\
    \hline
\end{tabular}
\label{tab:3}
\end{table*}

\begin{table*}[!t]
\caption{Predictive performances on the $15$ LDL datasets evaluated by \texttt{Clark distance} $\downarrow$.}
\centering
\scriptsize
\setlength{\tabcolsep}{6pt}
\begin{tabular}{ccccccccccc}
    \hline
    ID & LALOT & LDLLC & EDL-LRL & LDLSF & LDL-LCLR & LDL-SCL & LDL-LDM & LDL-LIFT-SAP\\
    \hline
    1  & 0.2187$\pm$0.0021 & 0.2163$\pm$0.0029 & 0.2157$\pm$0.0019 & 0.2110$\pm$0.0018 & 0.2114$\pm$0.0012 & 0.2101$\pm$0.0020 & \textbf{0.2099$\pm$0.0024} & 0.2103$\pm$0.0015 \\
    2  & 0.2217$\pm$0.0018 & 0.2205$\pm$0.0027 & 0.2216$\pm$0.0013 & 0.2167$\pm$0.0021 & 0.2170$\pm$0.0022 & 0.2165$\pm$0.0011 & \textbf{0.2164$\pm$0.0016} & \textbf{0.2164$\pm$0.0019} \\
    3  & 0.1491$\pm$0.0016 & 0.1464$\pm$0.0016 & 0.1456$\pm$0.0014 & 0.1403$\pm$0.0009 & 0.1405$\pm$0.0016 & \textbf{0.1391$\pm$0.0012} & 0.1402$\pm$0.0011 & 0.1399$\pm$0.0018 \\
    4  & 0.2244$\pm$0.0022 & 0.2200$\pm$0.0020 & 0.2209$\pm$0.0019 & 0.2018$\pm$0.0025 & 0.2024$\pm$0.0018 & 0.1993$\pm$0.0016 & 0.2018$\pm$0.0018 & \textbf{0.1990$\pm$0.0019} \\
    5  & 0.1034$\pm$0.0012 & 0.1004$\pm$0.0013 & 0.1002$\pm$0.0011 & 0.0986$\pm$0.0009 & \textbf{0.0982$\pm$0.0010} & 0.0986$\pm$0.0010 & 0.0987$\pm$0.0014 & 0.0983$\pm$0.0008 \\
    6  & 0.1901$\pm$0.0029 & 0.1837$\pm$0.0014 & 0.1848$\pm$0.0016 & 0.1838$\pm$0.0027 & \textbf{0.1829$\pm$0.0009} & 0.1831$\pm$0.0015 & 0.1831$\pm$0.0011 & 0.1831$\pm$0.0019 \\
    7  & 0.2602$\pm$0.0028 & 0.2573$\pm$0.0037 & 0.2559$\pm$0.0012 & 0.2519$\pm$0.0026 & 0.2511$\pm$0.0023 & \textbf{0.2475$\pm$0.0022} & 0.2509$\pm$0.0023 & 0.2506$\pm$0.0021 \\
    8  & 0.1856$\pm$0.0017 & 0.1864$\pm$0.0019 & 0.1840$\pm$0.0029 & 0.1850$\pm$0.0022 & 0.1843$\pm$0.0029 & \textbf{0.1828$\pm$0.0022} & 0.1843$\pm$0.0022 & 0.1850$\pm$0.0021 \\
    9 & 0.1315$\pm$0.0030 & 0.1298$\pm$0.0015 & 0.1297$\pm$0.0017 & 0.1300$\pm$0.0019 & 0.1297$\pm$0.0029 & 0.1296$\pm$0.0025 & \textbf{0.1291$\pm$0.0026} & 0.1301$\pm$0.0019 \\
    10 & 2.4845$\pm$0.0070 & 2.4979$\pm$0.0061 & 2.4952$\pm$0.0088 & 2.4615$\pm$0.0059 & \textbf{2.3599$\pm$0.0102} & 2.4594$\pm$0.0054 & 2.3976$\pm$0.0067 & 2.3616$\pm$0.0121 \\
    11 & 0.4267$\pm$0.0081 & 0.5071$\pm$0.0242 & 1.5582$\pm$0.0705 & 0.8482$\pm$0.2099 & 0.3995$\pm$0.0152 & 0.3458$\pm$0.0001 & 0.4187$\pm$0.0129 & \textbf{0.3384$\pm$0.0128} \\
    12 & 0.4129$\pm$0.0052 & 0.3663$\pm$0.0033 & 0.4259$\pm$0.0209 & 0.4273$\pm$0.0060 & 0.3816$\pm$0.0042 & 0.3696$\pm$0.0029 & 0.3665$\pm$0.0051 & \textbf{0.3326$\pm$0.0023} \\
    13 & 1.5686$\pm$0.0924 & 1.6640$\pm$0.0101 & 1.7782$\pm$0.0251 & 1.5380$\pm$0.0660 & 1.3209$\pm$0.0232 & 1.2075$\pm$0.0129 & 1.2970$\pm$0.0254 & \textbf{1.0784$\pm$0.0270} \\
    14 & 1.4855$\pm$0.0148 & 1.5134$\pm$0.0034 & 1.8385$\pm$0.0497 & 1.3881$\pm$0.0133 & 1.3907$\pm$0.0175 & 1.4335$\pm$0.0060 & 1.3898$\pm$0.0074 & \textbf{1.3777$\pm$0.0111} \\
    15 & 1.6956$\pm$0.0400 & 1.7119$\pm$0.0082 & 1.8028$\pm$0.0210 & 1.6821$\pm$0.0075 & 1.7189$\pm$0.0094 & \textbf{1.6510$\pm$0.0052} & 1.7068$\pm$0.0073 & 1.6782$\pm$0.0082 \\
    \hline
\end{tabular}
\label{tab:4}
\end{table*}

\begin{table*}[!t]
\caption{Predictive performances on the $15$ LDL data sets evaluated by \texttt{Canberra metric} $\downarrow$.}
\centering
\scriptsize
\setlength{\tabcolsep}{6pt}
\begin{tabular}{ccccccccccc}
    \hline
    ID & LALOT & LDLLC & EDL-LRL & LDLSF & LDL-LCLR & LDL-SCL & LDL-LDM & LDL-LIFT-SAP\\
    \hline
    1  & 0.7133$\pm$0.0071 & 0.7056$\pm$0.0095 & 0.7035$\pm$0.0057 & 0.6851$\pm$0.0054 & 0.6868$\pm$0.0032 & \textbf{0.6821$\pm$0.0052} & 0.6827$\pm$0.0067 & 0.6841$\pm$0.0051 \\
    2  & 0.6637$\pm$0.0065 & 0.6587$\pm$0.0085 & 0.6616$\pm$0.0037 & 0.6510$\pm$0.0046 & 0.6511$\pm$0.0058 & 0.6495$\pm$0.0031 & \textbf{0.6494$\pm$0.0046} & 0.6498$\pm$0.0053 \\
    3  & 0.2571$\pm$0.0030 & 0.2525$\pm$0.0028 & 0.2510$\pm$0.0026 & 0.2415$\pm$0.0015 & 0.2420$\pm$0.0026 & \textbf{0.2396$\pm$0.0021} & 0.2414$\pm$0.0019 & 0.2408$\pm$0.0032 \\
    4  & 0.4877$\pm$0.0057 & 0.4723$\pm$0.0041 & 0.4736$\pm$0.0036 & 0.4334$\pm$0.0053 & 0.4347$\pm$0.0032 & 0.4273$\pm$0.0035 & 0.4328$\pm$0.0035 & \textbf{0.4270$\pm$0.0040} \\
    5  & 0.1775$\pm$0.0019 & 0.1722$\pm$0.0022 & 0.1722$\pm$0.0019 & 0.1696$\pm$0.0015 & 0.1691$\pm$0.0016 & 0.1694$\pm$0.0018 & 0.1697$\pm$0.0022 & \textbf{0.1690$\pm$0.0015} \\
    6  & 0.3800$\pm$0.0056 & 0.3662$\pm$0.0024 & 0.3685$\pm$0.0034 & 0.3659$\pm$0.0048 & \textbf{0.3642$\pm$0.0017} & 0.3650$\pm$0.0030 & 0.3652$\pm$0.0024 & 0.3652$\pm$0.0032 \\
    7  & 0.5380$\pm$0.0052 & 0.5312$\pm$0.0075 & 0.5281$\pm$0.0024 & 0.5173$\pm$0.0057 & 0.5161$\pm$0.0053 & \textbf{0.5087$\pm$0.0048} & 0.5160$\pm$0.0044 & 0.5151$\pm$0.0041 \\
    8  & 0.2853$\pm$0.0024 & 0.2862$\pm$0.0031 & 0.2826$\pm$0.0044 & 0.2844$\pm$0.0032 & 0.2834$\pm$0.0045 & \textbf{0.2809$\pm$0.0034} & 0.2830$\pm$0.0033 & 0.2842$\pm$0.0028 \\
    9 & 0.1832$\pm$0.0041 & 0.1807$\pm$0.0020 & 0.1806$\pm$0.0024 & 0.1808$\pm$0.0026 & 0.1804$\pm$0.0039 & 0.1804$\pm$0.0035 & \textbf{0.1796$\pm$0.0035} & 0.1809$\pm$0.0026 \\
    10 & 6.9970$\pm$0.0302 & 7.0048$\pm$0.0222 & 7.0507$\pm$0.0370 & 6.7846$\pm$0.0231 & 6.5179$\pm$0.0355 & 6.7446$\pm$0.0208 & 6.6010$\pm$0.0265 & \textbf{6.4065$\pm$0.0486} \\
    11 & 0.8953$\pm$0.0155 & 1.0452$\pm$0.0536 & 3.4685$\pm$0.1797 & 1.6643$\pm$0.4082 & 0.8195$\pm$0.0310 & 0.7157$\pm$0.0007 & 0.8605$\pm$0.0282 & \textbf{0.7011$\pm$0.0279} \\
    12 & 0.8974$\pm$0.0096 & 0.7653$\pm$0.0066 & 0.8860$\pm$0.0440 & 0.8694$\pm$0.0100 & 0.7916$\pm$0.0083 & 0.7862$\pm$0.0069 & 0.7624$\pm$0.0108 & \textbf{0.6924$\pm$0.0044} \\
    13 & 3.1952$\pm$0.2598 & 3.4733$\pm$0.0278 & 3.7189$\pm$0.0634 & 3.0393$\pm$0.1846 & 2.5953$\pm$0.0548 & 2.3147$\pm$0.0260 & 2.5480$\pm$0.0579 & \textbf{2.0489$\pm$0.0501} \\
    14 & 2.9123$\pm$0.0503 & 2.9866$\pm$0.0100 & 3.8475$\pm$0.1152 & 2.5794$\pm$0.0234 & 2.6020$\pm$0.0464 & 2.7458$\pm$0.0086 & 2.5952$\pm$0.0211 & \textbf{2.5564$\pm$0.0266} \\
    15 & 3.8727$\pm$0.1064 & 3.9067$\pm$0.0235 & 4.2042$\pm$0.0680 & 3.7494$\pm$0.0215 & 3.9054$\pm$0.0278 & \textbf{3.7014$\pm$0.0148} & 3.8682$\pm$0.0226 & 3.7720$\pm$0.0181 \\
    \hline
\end{tabular}
\label{tab:5}
\end{table*}

\begin{table*}[!t]
\caption{Predictive performances on the $15$ LDL data sets evaluated by \texttt{K-L divergence} $\downarrow$.}
\centering
\scriptsize
\setlength{\tabcolsep}{6pt}
\begin{tabular}{ccccccccccc}
    \hline
    ID & LALOT & LDLLC & EDL-LRL & LDLSF & LDL-LCLR & LDL-SCL & LDL-LDM & LDL-LIFT-SAP\\
    \hline
    1  & 0.0059$\pm$0.0001 & 0.0058$\pm$0.0001 & 0.0058$\pm$0.0001 & 0.0056$\pm$0.0001 & 0.0056$\pm$0.0001 & \textbf{0.0055$\pm$0.0001} & \textbf{0.0055$\pm$0.0001} & \textbf{0.0055$\pm$0.0001} \\
    2  & 0.0075$\pm$0.0001 & 0.0074$\pm$0.0002 & 0.0074$\pm$0.0001 & \textbf{0.0070$\pm$0.0001} & \textbf{0.0070$\pm$0.0001} & \textbf{0.0070$\pm$0.0001} & \textbf{0.0070$\pm$0.0001} & \textbf{0.0070$\pm$0.0001} \\
    3  & 0.0139$\pm$0.0003 & 0.0135$\pm$0.0003 & 0.0133$\pm$0.0003 & 0.0123$\pm$0.0002 & 0.0123$\pm$0.0003 & \textbf{0.0122$\pm$0.0002} & 0.0123$\pm$0.0002 & \textbf{0.0122$\pm$0.0003} \\
    4  & 0.0160$\pm$0.0003 & 0.0155$\pm$0.0003 & 0.0156$\pm$0.0003 & 0.0133$\pm$0.0003 & 0.0134$\pm$0.0003 & \textbf{0.0129$\pm$0.0002} & 0.0132$\pm$0.0002 & 0.0130$\pm$0.0003 \\
    5  & 0.0068$\pm$0.0002 & 0.0065$\pm$0.0002 & 0.0065$\pm$0.0001 & 0.0063$\pm$0.0002 & \textbf{0.0062$\pm$0.0002} & 0.0063$\pm$0.0002 & 0.0063$\pm$0.0002 & \textbf{0.0062$\pm$0.0001} \\
    6  & 0.0138$\pm$0.0004 & 0.0128$\pm$0.0002 & 0.0130$\pm$0.0003 & 0.0128$\pm$0.0003 & \textbf{0.0127$\pm$0.0001} & \textbf{0.0127$\pm$0.0002} & \textbf{0.0127$\pm$0.0001} & \textbf{0.0127$\pm$0.0003} \\
    7  & 0.0281$\pm$0.0007 & 0.0273$\pm$0.0008 & 0.0270$\pm$0.0003 & 0.0250$\pm$0.0004 & 0.0249$\pm$0.0005 & \textbf{0.0242$\pm$0.0005} & 0.0248$\pm$0.0004 & 0.0248$\pm$0.0004 \\
    8  & 0.0301$\pm$0.0005 & 0.0301$\pm$0.0006 & 0.0293$\pm$0.0009 & 0.0295$\pm$0.0008 & 0.0294$\pm$0.0007 & \textbf{0.0290$\pm$0.0007} & 0.0293$\pm$0.0006 & 0.0295$\pm$0.0006 \\
    9 & 0.0270$\pm$0.0012 & 0.0262$\pm$0.0006 & 0.0261$\pm$0.0007 & 0.0247$\pm$0.0008 & 0.0247$\pm$0.0011 & 0.0247$\pm$0.0008 & \textbf{0.0246$\pm$0.0009} & 0.0248$\pm$0.0007 \\
    10 & 1.1637$\pm$0.0169 & 1.3696$\pm$0.0594 & 3.1072$\pm$0.3137 & 1.4385$\pm$0.0474 & 1.2583$\pm$0.0881 & 0.8290$\pm$0.0111 & 0.9094$\pm$0.0494 & \textbf{0.6855$\pm$0.0117} \\
    11 & 0.0730$\pm$0.0025 & 0.1033$\pm$0.0104 & 1.3538$\pm$0.2548 & 0.6391$\pm$0.3919 & 0.0668$\pm$0.0056 & 0.0491$\pm$0.0000 & 0.0715$\pm$0.0046 & \textbf{0.0482$\pm$0.0044} \\
    12 & 0.0858$\pm$0.0029 & 0.0593$\pm$0.0009 & 0.0996$\pm$0.0109 & 0.0840$\pm$0.0045 & 0.0640$\pm$0.0015 & 0.0625$\pm$0.0009 & 0.0595$\pm$0.0018 & \textbf{0.0525$\pm$0.0008} \\
    13 & 0.8174$\pm$0.0922 & 2.0968$\pm$0.0989 & 5.2367$\pm$0.3383 & 1.3218$\pm$0.0830 & 1.0771$\pm$0.0480 & \textbf{0.5031$\pm$0.0117} & 1.0601$\pm$0.0878 & 0.5641$\pm$0.0283 \\
    14 & 0.6582$\pm$0.0225 & 0.8105$\pm$0.0227 & 49.7839$\pm$19.4004 & 1.2984$\pm$0.0821 & 0.4648$\pm$0.0806 & 0.5200$\pm$0.0054 & 0.4200$\pm$0.0275 & \textbf{0.3712$\pm$0.0298} \\
    15 & 0.7243$\pm$0.0559 & 0.7358$\pm$0.0154 & 1.3083$\pm$0.2251 & 0.8927$\pm$0.0274 & 0.6562$\pm$0.0136 & \textbf{0.5775$\pm$0.0045} & 0.6356$\pm$0.0105 & 0.6042$\pm$0.0068 \\
    \hline
\end{tabular}
\label{tab:6}
\end{table*}

\begin{table*}[!t]
\caption{Predictive performances on the $15$ LDL data sets evaluated by \texttt{Cosine coefficient} $\uparrow$.}
\centering
\scriptsize
\setlength{\tabcolsep}{6pt}
\begin{tabular}{ccccccccccc}
    \hline
    ID & LALOT & LDLLC & EDL-LRL & LDLSF & LDL-LCLR & LDL-SCL & LDL-LDM & LDL-LIFT-SAP\\
    \hline
    1  & 0.9942$\pm$0.0001 & 0.9943$\pm$0.0001 & 0.9944$\pm$0.0001 & \textbf{0.9946$\pm$0.0001} & 0.9945$\pm$0.0001 & \textbf{0.9946$\pm$0.0001} & \textbf{0.9946$\pm$0.0001} & \textbf{0.9946$\pm$0.0001} \\
    2  & 0.9928$\pm$0.0001 & 0.9929$\pm$0.0002 & 0.9929$\pm$0.0001 & 0.9932$\pm$0.0001 & 0.9932$\pm$0.0001 & \textbf{0.9933$\pm$0.0001} & \textbf{0.9933$\pm$0.0001} & \textbf{0.9933$\pm$0.0001} \\
    3  & 0.9869$\pm$0.0003 & 0.9873$\pm$0.0003 & 0.9875$\pm$0.0002 & 0.9884$\pm$0.0002 & 0.9884$\pm$0.0003 & \textbf{0.9886$\pm$0.0002} & 0.9884$\pm$0.0001 & 0.9885$\pm$0.0003 \\
    4  & 0.9851$\pm$0.0003 & 0.9857$\pm$0.0002 & 0.9856$\pm$0.0002 & 0.9878$\pm$0.0003 & 0.9877$\pm$0.0002 & \textbf{0.9881$\pm$0.0002} & 0.9878$\pm$0.0001 & 0.9880$\pm$0.0002 \\
    5  & 0.9935$\pm$0.0001 & 0.9939$\pm$0.0002 & 0.9939$\pm$0.0001 & 0.9940$\pm$0.0001 & \textbf{0.9941$\pm$0.0001} & 0.9940$\pm$0.0001 & 0.9940$\pm$0.0002 & \textbf{0.9941$\pm$0.0001} \\
    6  & 0.9869$\pm$0.0004 & 0.9878$\pm$0.0002 & 0.9877$\pm$0.0002 & 0.9879$\pm$0.0003 & \textbf{0.9880$\pm$0.0001} & 0.9879$\pm$0.0002 & 0.9879$\pm$0.0002 & \textbf{0.9880$\pm$0.0002} \\
    7  & 0.9737$\pm$0.0006 & 0.9745$\pm$0.0007 & 0.9748$\pm$0.0003 & 0.9766$\pm$0.0004 & 0.9767$\pm$0.0004 & \textbf{0.9773$\pm$0.0005} & 0.9767$\pm$0.0004 & 0.9767$\pm$0.0004 \\
    8  & 0.9735$\pm$0.0004 & 0.9735$\pm$0.0006 & 0.9741$\pm$0.0007 & 0.9738$\pm$0.0006 & 0.9740$\pm$0.0006 & \textbf{0.9744$\pm$0.0005} & 0.9740$\pm$0.0005 & 0.9739$\pm$0.0005 \\
    9 & 0.9770$\pm$0.0009 & 0.9776$\pm$0.0004 & 0.9777$\pm$0.0005 & 0.9788$\pm$0.0006 & 0.9788$\pm$0.0007 & 0.9789$\pm$0.0006 & \textbf{0.9790$\pm$0.0006} & 0.9787$\pm$0.0006 \\
    10 & 0.5790$\pm$0.0060 & 0.6251$\pm$0.0063 & 0.5896$\pm$0.0109 & 0.6939$\pm$0.0038 & 0.6372$\pm$0.0122 & 0.7174$\pm$0.0039 & 0.7025$\pm$0.0127 & \textbf{0.7665$\pm$0.0063} \\
    11 & 0.9309$\pm$0.0026 & 0.9081$\pm$0.0090 & 0.5990$\pm$0.0256 & 0.8610$\pm$0.0409 & 0.9392$\pm$0.0048 & 0.9534$\pm$0.0000 & 0.9358$\pm$0.0047 & \textbf{0.9547$\pm$0.0043} \\
    12 & 0.9172$\pm$0.0024 & 0.9429$\pm$0.0008 & 0.9318$\pm$0.0057 & 0.9351$\pm$0.0009 & 0.9384$\pm$0.0015 & 0.9382$\pm$0.0009 & 0.9422$\pm$0.0015 & \textbf{0.9482$\pm$0.0006} \\
    13 & 0.6306$\pm$0.0532 & 0.4647$\pm$0.0155 & 0.4437$\pm$0.0106 & 0.6978$\pm$0.0137 & 0.6508$\pm$0.0115 & \textbf{0.7682$\pm$0.0047} & 0.6626$\pm$0.0136 & 0.7531$\pm$0.0086 \\
    14 & 0.6850$\pm$0.0142 & 0.6523$\pm$0.0042 & 0.5005$\pm$0.0400 & 0.7776$\pm$0.0056 & 0.8265$\pm$0.0134 & 0.7749$\pm$0.0038 & 0.8332$\pm$0.0052 & \textbf{0.8466$\pm$0.0047} \\
    15 & 0.6549$\pm$0.0098 & 0.6566$\pm$0.0057 & 0.6348$\pm$0.0072 & 0.6985$\pm$0.0054 & 0.6907$\pm$0.0059 & \textbf{0.7208$\pm$0.0018} & 0.6985$\pm$0.0042 & 0.7058$\pm$0.0028 \\
    \hline
\end{tabular}
\label{tab:7}
\end{table*}

\begin{table*}[!t]
\caption{Predictive performances on the $15$ LDL datasets evaluated by \texttt{Intersection similarity} $\uparrow$.}
\centering
\scriptsize
\setlength{\tabcolsep}{6pt}
\begin{tabular}{ccccccccccc}
    \hline
    ID & LALOT & LDLLC & EDL-LRL & LDLSF & LDL-LCLR & LDL-SCL & LDL-LDM & LDL-LIFT-SAP\\
    \hline
    1  & 0.9608$\pm$0.0004 & 0.9610$\pm$0.0005 & 0.9612$\pm$0.0003 & \textbf{0.9624$\pm$0.0003} & 0.9622$\pm$0.0002 & \textbf{0.9624$\pm$0.0003} & 0.9623$\pm$0.0004 & 0.9622$\pm$0.0003 \\
    2  & 0.9564$\pm$0.0004 & 0.9566$\pm$0.0006 & 0.9565$\pm$0.0002 & \textbf{0.9573$\pm$0.0003} & \textbf{0.9573$\pm$0.0004} & \textbf{0.9573$\pm$0.0002} & \textbf{0.9573$\pm$0.0003} & \textbf{0.9573$\pm$0.0003} \\
    3  & 0.9365$\pm$0.0008 & 0.9377$\pm$0.0007 & 0.9381$\pm$0.0006 & 0.9405$\pm$0.0004 & 0.9404$\pm$0.0006 & \textbf{0.9410$\pm$0.0005} & 0.9405$\pm$0.0005 & 0.9407$\pm$0.0008 \\
    4  & 0.9321$\pm$0.0008 & 0.9342$\pm$0.0005 & 0.9341$\pm$0.0005 & 0.9400$\pm$0.0008 & 0.9398$\pm$0.0004 & \textbf{0.9408$\pm$0.0005} & 0.9400$\pm$0.0005 & \textbf{0.9408$\pm$0.0005} \\
    5  & 0.9562$\pm$0.0005 & 0.9575$\pm$0.0005 & 0.9575$\pm$0.0005 & 0.9582$\pm$0.0004 & \textbf{0.9583$\pm$0.0003} & 0.9582$\pm$0.0004 & 0.9581$\pm$0.0005 & \textbf{0.9583$\pm$0.0004} \\
    6  & 0.9376$\pm$0.0009 & 0.9399$\pm$0.0004 & 0.9395$\pm$0.0006 & 0.9400$\pm$0.0008 & \textbf{0.9403$\pm$0.0003} & 0.9401$\pm$0.0005 & 0.9401$\pm$0.0004 & 0.9401$\pm$0.0005 \\
    7  & 0.9113$\pm$0.0009 & 0.9125$\pm$0.0012 & 0.9130$\pm$0.0004 & 0.9149$\pm$0.0009 & 0.9151$\pm$0.0009 & \textbf{0.9162$\pm$0.0008} & 0.9150$\pm$0.0007 & 0.9152$\pm$0.0007 \\
    8  & 0.9078$\pm$0.0008 & 0.9076$\pm$0.0011 & 0.9087$\pm$0.0014 & 0.9081$\pm$0.0010 & 0.9084$\pm$0.0014 & \textbf{0.9092$\pm$0.0010} & 0.9085$\pm$0.0010 & 0.9082$\pm$0.0008 \\
    9 & 0.9113$\pm$0.0019 & 0.9125$\pm$0.0009 & 0.9126$\pm$0.0011 & 0.9127$\pm$0.0012 & 0.9128$\pm$0.0018 & 0.9129$\pm$0.0017 & \textbf{0.9132$\pm$0.0016} & 0.9126$\pm$0.0012 \\
    10 & 0.3651$\pm$0.0047 & 0.4715$\pm$0.0034 & 0.3999$\pm$0.0097 & 0.5254$\pm$0.0037 & 0.5024$\pm$0.0098 & 0.5036$\pm$0.0026 & 0.5500$\pm$0.0092 & \textbf{0.6087$\pm$0.0055} \\
    11 & 0.8473$\pm$0.0031 & 0.8234$\pm$0.0095 & 0.4752$\pm$0.0254 & 0.7618$\pm$0.0488 & 0.8620$\pm$0.0056 & 0.8779$\pm$0.0000 & 0.8557$\pm$0.0055 & \textbf{0.8812$\pm$0.0052} \\
    12 & 0.8392$\pm$0.0017 & 0.8652$\pm$0.0010 & 0.8420$\pm$0.0075 & 0.8527$\pm$0.0012 & 0.8604$\pm$0.0016 & 0.8588$\pm$0.0012 & 0.8650$\pm$0.0018 & \textbf{0.8759$\pm$0.0006} \\
    13 & 0.4522$\pm$0.0386 & 0.3781$\pm$0.0107 & 0.3532$\pm$0.0098 & 0.5702$\pm$0.0115 & 0.5552$\pm$0.0088 & 0.6131$\pm$0.0024 & 0.5650$\pm$0.0117 & \textbf{0.6342$\pm$0.0060} \\
    14 & 0.5200$\pm$0.0129 & 0.5033$\pm$0.0031 & 0.3460$\pm$0.0130 & 0.6430$\pm$0.0035 & 0.6867$\pm$0.0099 & 0.6186$\pm$0.0027 & 0.6918$\pm$0.0038 & \textbf{0.7073$\pm$0.0037} \\
    15 & 0.5273$\pm$0.0064 & 0.5421$\pm$0.0034 & 0.4962$\pm$0.0122 & 0.5747$\pm$0.0036 & 0.5738$\pm$0.0037 & 0.5794$\pm$0.0018 & 0.5786$\pm$0.0039 & \textbf{0.5836$\pm$0.0027} \\
    \hline
\end{tabular}
\label{tab:8}
\end{table*}

Tabs.~\ref{tab:3}--\ref{tab:8} present the predictive performance of each comparison algorithm w.r.t. \emph{Chebyshev distance}, \emph{Clark distance}, \emph{Canberra metric}, \emph{Kullback-Leibler divergence}, \emph{Cosine coefficient}, and \emph{Intersection similarity}, respectively. From these tables, it is evident that, overall, LDL-LIFT-SAP ranks first or is tied for first in $55.56\%$ of the $90$ cases ($15$ datasets $\times$ $6$ evaluation metrics) and secures second place or ties for second in $24.44\%$ of cases. These findings clearly show the superior predictive performance of LDL-LIFT-SAP.

Meanwhile, we observe that on the first nine datasets (bioinformatics datasets), LDL-LIFT-SAP achieves the best or tied-for-best results in only $38.89\%$ (21 out of 54) of cases. In contrast, on the last six datasets (image datasets), it demonstrates the most competitive performance in $80.56\%$ (29 out of 36) of cases. Notably, compared to seven state-of-the-art LDL algorithms, LDL-LIFT-SAP delivers significant improvements on \emph{Natural Scene} and \emph{M$^2$B} datasets for \emph{Chebyshev distance}; on \emph{S-BU\_3DFE} and \emph{M$^2$B} datasets for \emph{Clark distance} and \emph{Canberra metric}; on \emph{Natural Scene}, \emph{S-BU\_3DFE}, and \emph{SCUT-FBP} datasets for \emph{Kullback-Leibler divergence}; and on \emph{Natural Scene} dataset for \emph{Cosine coefficient} and \emph{Intersection similarity}. These results indicate that LDL-LIFT-SAP performs more effectively on LDL tasks with dense feature representations than on those with sparse features.

\subsection{Statistical Tests}
To provide a more rigorous performance analysis, the Friedman test \cite{friedman:test}, a widely used statistical method for comparing multiple algorithms across various datasets, is conducted. Suppose there are $s$ algorithms to be compared and $N$ data sets. Let $r_i^j$ represent the ranking of the $j$-th algorithm on the $i$-th data set, and $R_j = \frac{1}{N}\sum_{i=1}^{N}r_i^j$ be the average ranking of the $j$-th algorithm across all the data sets. 

Under the null hypothesis, the Friedman statistic $F_F$ follows a Fisher distribution with $(s-1)$ and $(s-1)(N-1)$ degrees of freedom:

\begin{equation}
F_F = \frac{(N-1)\chi_F^2}{N(s-1)-\chi_F^2},
\label{eq:15}
\end{equation}
where
\begin{equation}
\chi_F^2 = \frac{12N}{s(s+1)}(\sum_{j=1}^{s}R_j^2-\frac{s(s+1)^2}{4}).
\label{eq:16}
\end{equation}

\begin{table}[!t]
\caption{Friedman statistic and critical value.}
\centering
\label{tab:9}
\begin{tabular}{ccccc}
\hline
Evaluation metrics & $F_F$ & CV ($\alpha = 0.05$) \\
\hline
 \multicolumn{1}{c}{\emph{Chebyshev distance}} &  \multicolumn{1}{c}{$24.2646$} & \multirow{6}{*}{$2.1044$}\\
\emph{Clark distance}           &        $19.5191$         &                   \\
\emph{Canberra metric}          &        $26.2556$         &                   \\
\emph{Kullback-Leibler divergence}&        $25.1246$       &                   \\
\emph{Cosine coefficient}       &        $31.1767$         &                   \\
\emph{Intersection similarity}  &        $27.2471$         &                   \\
\hline
\end{tabular}
\end{table}

Tab.~\ref{tab:9} shows the Friedman statistic $F_F$ ($s = 8$, $N = 15$) and the critical value (CV). From Tab.~\ref{tab:9}, we can observe that at a significance level of $0.05$, the null hypothesis of `equal performance' among all eight algorithms is decisively rejected for each evaluation metric. Consequently, the \emph{Nemenyi test} is conducted to evaluate whether the proposed LDL-LIFT-SAP algorithm exhibits a statistically significant difference in predictive performance compared to the other seven LDL algorithms, using LDL-LIFT-SAP as the control algorithm. A significant difference between two algorithms is identified if their average rankings over all datasets differ by at least one critical difference (CD):

\begin{equation}
CD = q_\alpha \sqrt{\frac{s(s+1)}{6N}},
\label{eq:17}
\end{equation}
where the critical value $q_\alpha$ is based on the studentized range statistic divided by $\sqrt{2}$. At a significance level of $\alpha = 0.05$, we have $q_\alpha = 3.0310$, and then $CD = 2.7110$.

\begin{figure}[!t]
\centering
\subfloat[Chebyshev distance]{\includegraphics[width=0.23\textwidth]{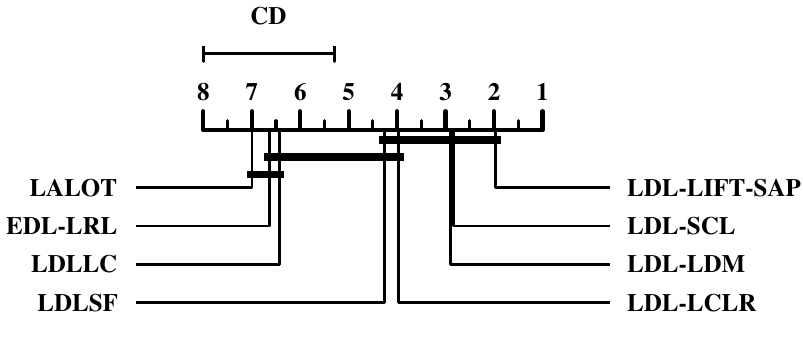}
\label{fig:4a}}
\hfil
\subfloat[Clark distance]{\includegraphics[width=0.23\textwidth]{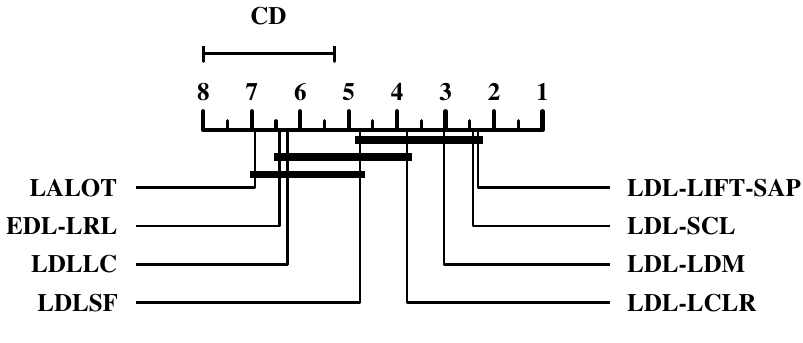}
\label{fig:4b}}\\
\subfloat[Canberra metric]{\includegraphics[width=0.23\textwidth]{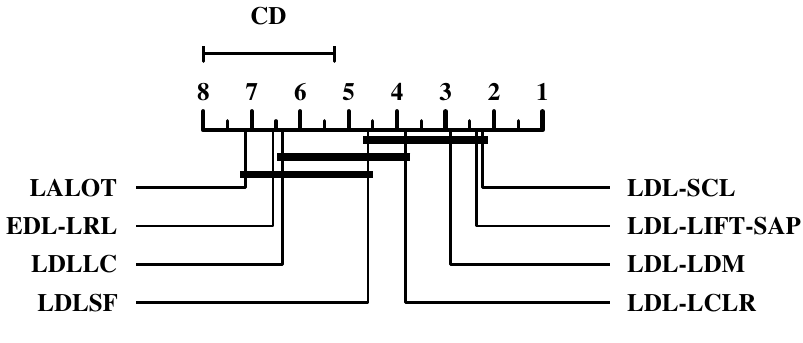}
\label{fig:4c}}
\hfil
\subfloat[K-L divergence]{\includegraphics[width=0.23\textwidth]{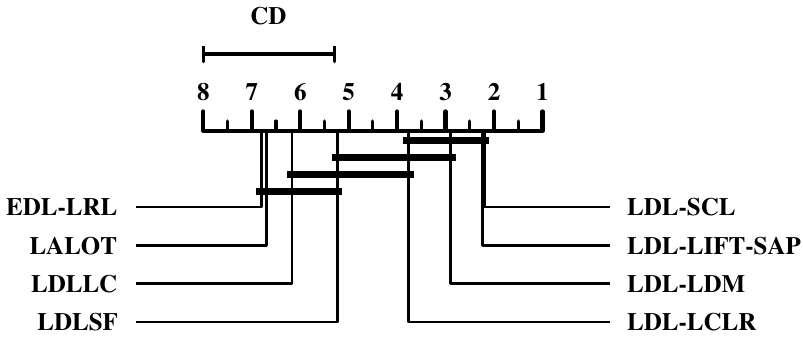}
\label{fig:4d}}\\
\subfloat[Cosine coefficient]{\includegraphics[width=0.23\textwidth]{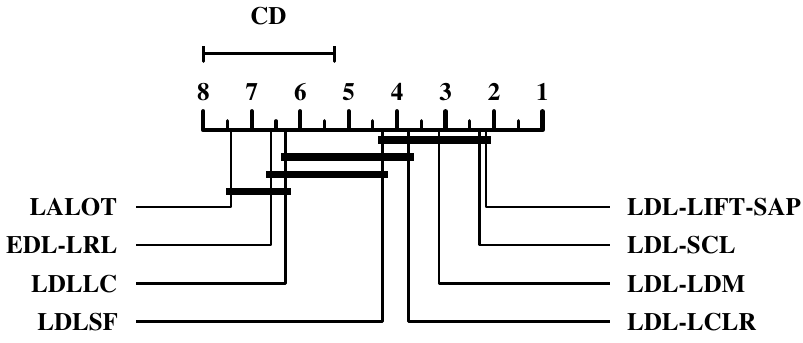}
\label{fig:4e}}
\hfil
\subfloat[Intersection similarity]{\includegraphics[width=0.23\textwidth]{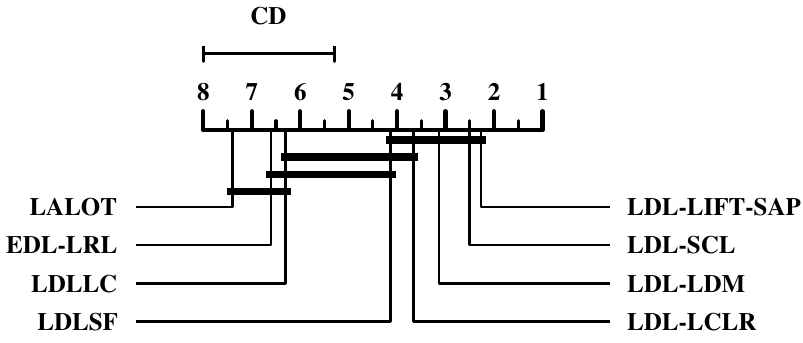}
\label{fig:4f}}
\caption{CD diagrams on the six evaluation metrics.}
\label{fig:4}
\end{figure}

To visually illustrate the actual differences in predictive performances between LDL-LIFT-SAP and the other seven algorithms, Fig.~\ref{fig:4} presents the CD diagrams for each evaluation metric. In each subfigure, the average ranking of each LDL algorithm is marked along the axis, with lower ranks appearing to the right. Algorithms that are not significantly different from each other according to the \emph{Nemenyi test} are connected by a thick horizontal line. The value of the CD is also displayed above the corresponding axis.

As shown in Fig.~\ref{fig:4}, among the $42$ pairwise comparisons ($7$ competing algorithms $\times$ $6$ evaluation metrics), LDL-LIFT-SAP achieves statistically comparable performance in $54.76\%$ of the cases and statistically superior performance in $45.24\%$ of the cases, confirming its overall competitive advantage.

From Fig.5(a) to~5(f), we observe that LDL-LIFT-SAP consistently ranks among the top two positions under all six evaluation metrics:
\begin{itemize}
    \item Under the \emph{Chebyshev distance} (Fig.5(a)), \emph{Clark distance} (Fig.5(b)), \emph{Cosine coefficient} (Fig.5(e)) and \emph{Intersection similarity} (Fig.5(f)), LDL-LIFT-SAP is ranked first, with a statistically significant margin over most other LDL algorithms.
    \item For the \emph{Canberra metric} (Fig.5(c)) and \emph{Kullback-Leibler divergence} (Fig.5(d)), LDL-SCL achieves the highest rank, while LDL-LIFT-SAP follows closely as the second-best performer, exhibiting performance that is statistically comparable to LDL-SCL.
\end{itemize}

These results demonstrate the statistical superiority of LDL-LIFT-SAP across diverse distance- and similarity-based metrics. It consistently outperforms or matches the performance of seven well-established LDL algorithms, validating its effectiveness as a state-of-the-art solution for LDL tasks.

\subsection{Ablation Analysis}

\begin{figure}[!t]
\centering
\subfloat[Chebyshev distance]{\includegraphics[width=0.23\textwidth]{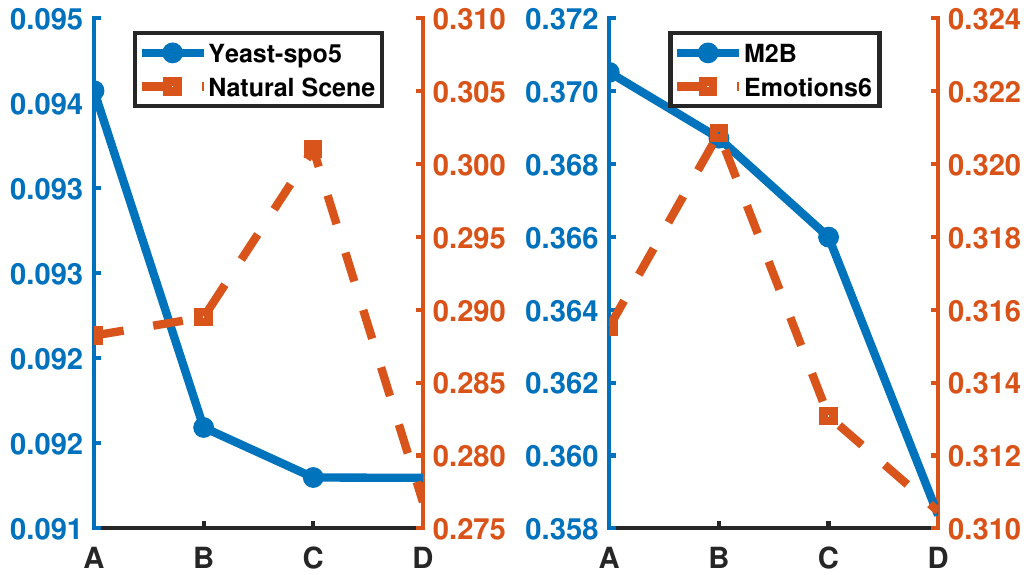}
\label{fig:5a}}
\hfil
\subfloat[Clark distance]{\includegraphics[width=0.23\textwidth]{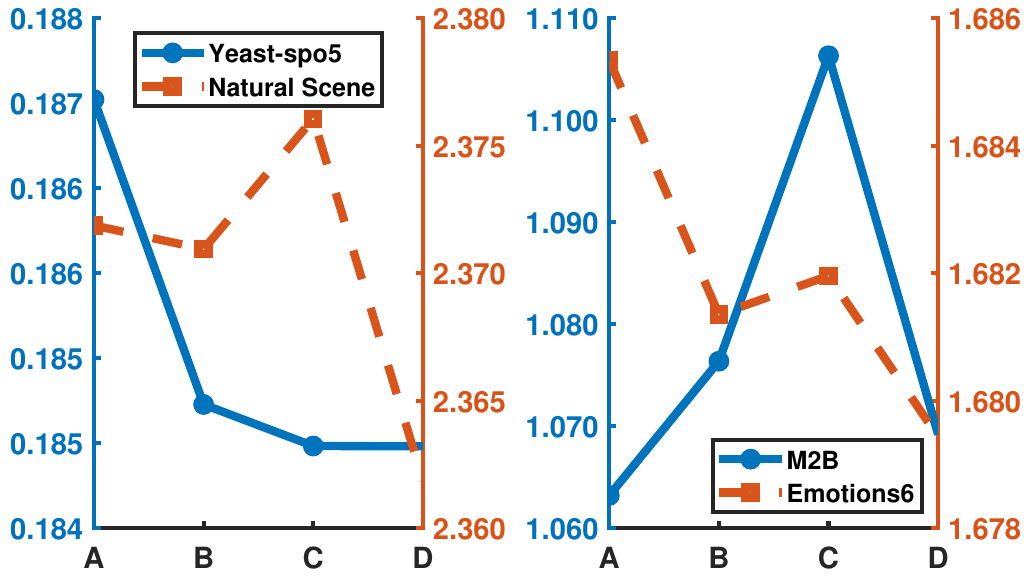}
\label{fig:5b}}\\
\subfloat[Canberra metric]{\includegraphics[width=0.23\textwidth]{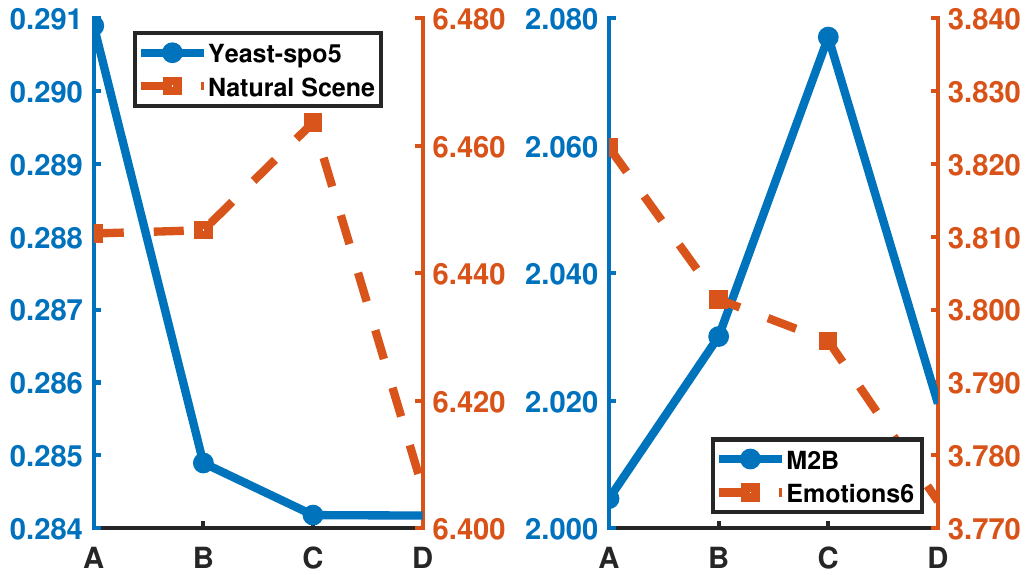}
\label{fig:5c}}
\hfil
\subfloat[K-L divergence]{\includegraphics[width=0.23\textwidth]{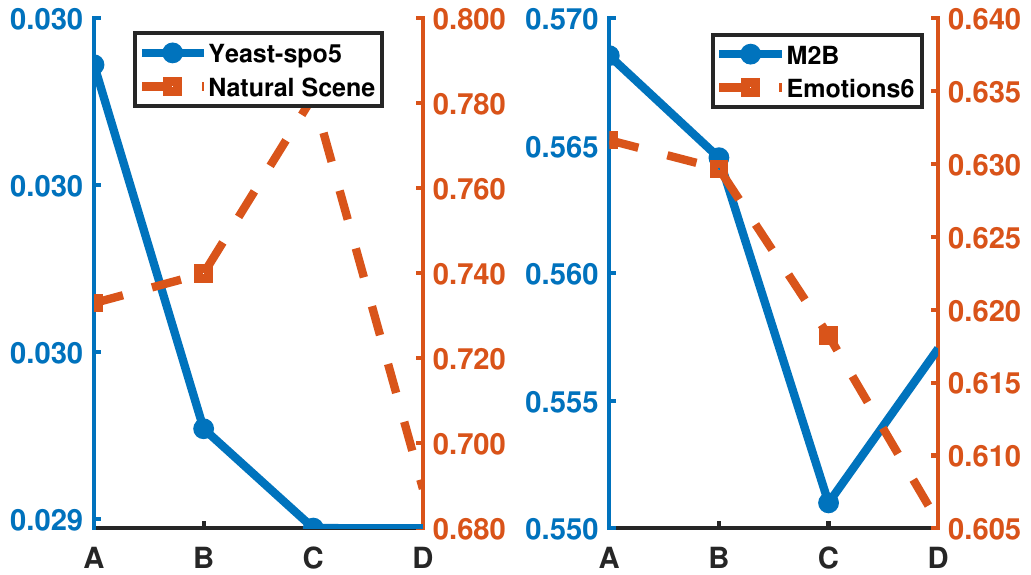}
\label{fig:5d}}\\
\subfloat[Cosine coefficient]{\includegraphics[width=0.23\textwidth]{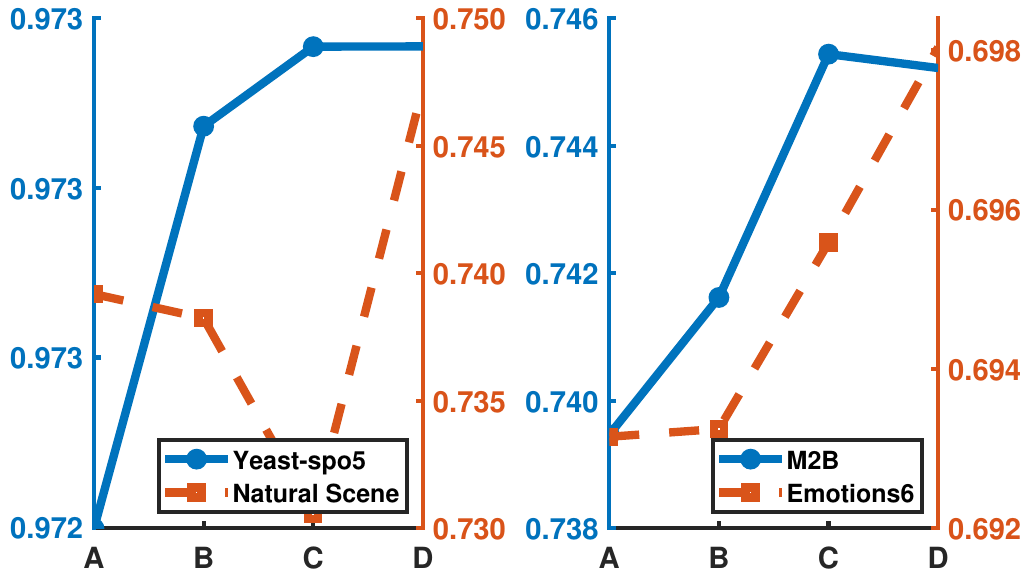}
\label{fig:5e}}
\hfil
\subfloat[Intersection similarity]{\includegraphics[width=0.23\textwidth]{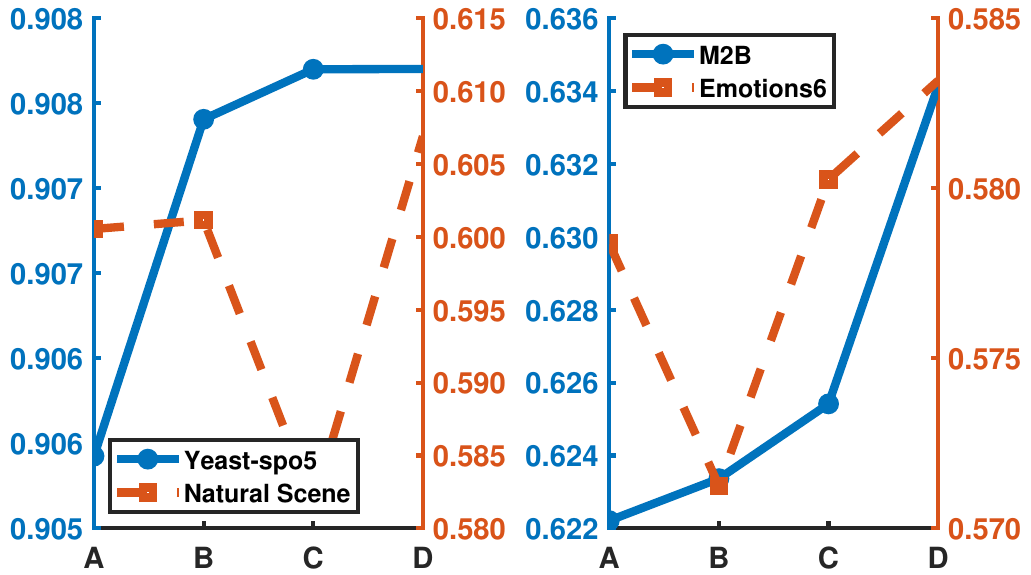}
\label{fig:5f}}
\caption{Ablation analysis on the datasets from four distinct domains.}
\label{fig:5}
\end{figure}

LIFT-SAP re-characterizes each instance by integrating distance and directional information relative to SAPs into the LIFT framework, significantly enhancing the discrimination of the feature space for each label and ultimately enabling the superior performance of LDL-LIFT-SAP. Fig.\ref{fig:5} shows the performance of LIFT-SAP and its three variants: `A' (LIFT), `B' (LIFT-SAP without directional information relative to SAPs), `C' (LIFT-SAP without distance information relative to SAPs), and `D' (LIFT-SAP), across four datasets, i.e., \emph{Yeast-spo5}, \emph{Natural Scene}, \emph{M$^2$B}, and \emph{Emotions6}, from different domains.

Ablation experiments show that introducing SAPs significantly enhances LIFT's characterization capabilities by capturing interactions among distinct clusters, leading to notable improvements in nearly all cases (except for \emph{M$^2$B} dataset on \emph{Clark distance} and \emph{Canberra metric}). Additionally, LIFT-SAP generally outperforms its two variants that rely solely on either distance or directional information relative to SAPs, highlighting the complementary nature of two perspectives in uncovering each label's unique characteristics and enhancing feature representations. These findings validate the rationality behind LIFT-SAP and its effectiveness in LDL tasks.

\subsection{Parameter Sensitivity Analysis}
In this analysis, we focus on evaluating the impact of the parameters $\lambda$, $\mu$, and $\varepsilon$ in the serial fusion scheme defined in Eq.~\eqref{eq:13}. The ratio parameter $\sigma$ and the discount factor $\alpha$ are directly inherited from LIFT~\cite{zhang-wu:lift} and UCL~\cite{dong-et-al:label-specific}, and are therefore not the focus of our investigation in LIFT-SAP.

\begin{figure}[!t]
\centering
\subfloat[Chebyshev distance]{\includegraphics[width=0.23\textwidth]{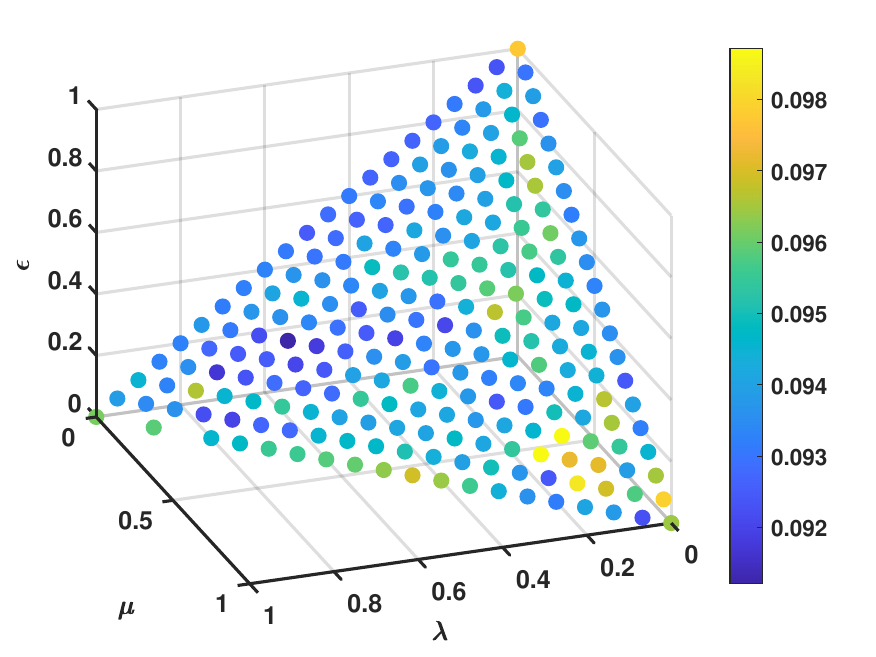}
\label{fig:6a}}
\hfil
\subfloat[Clark distance]{\includegraphics[width=0.23\textwidth]{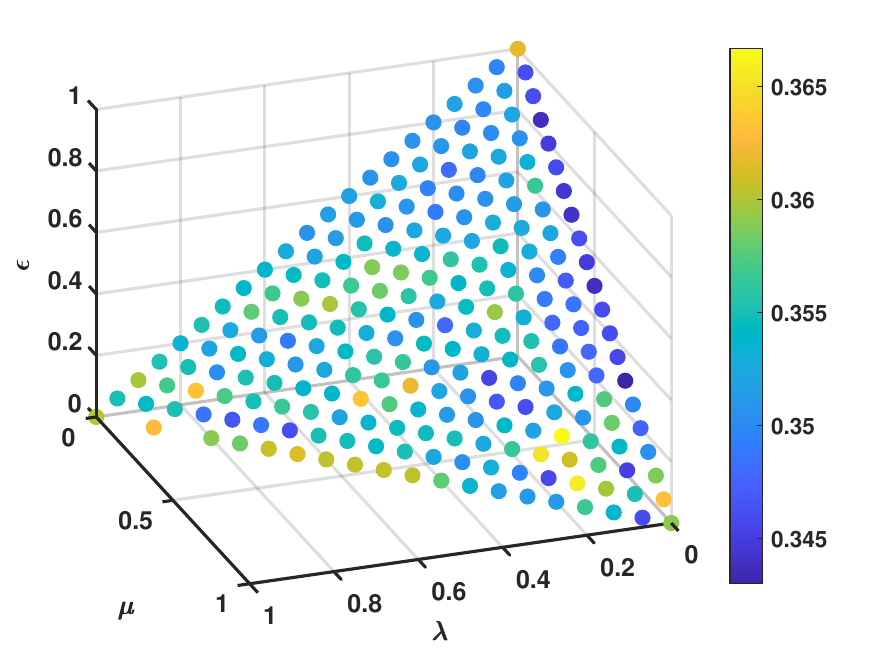}
\label{fig:6b}}\\
\subfloat[Canberra metric]{\includegraphics[width=0.23\textwidth]{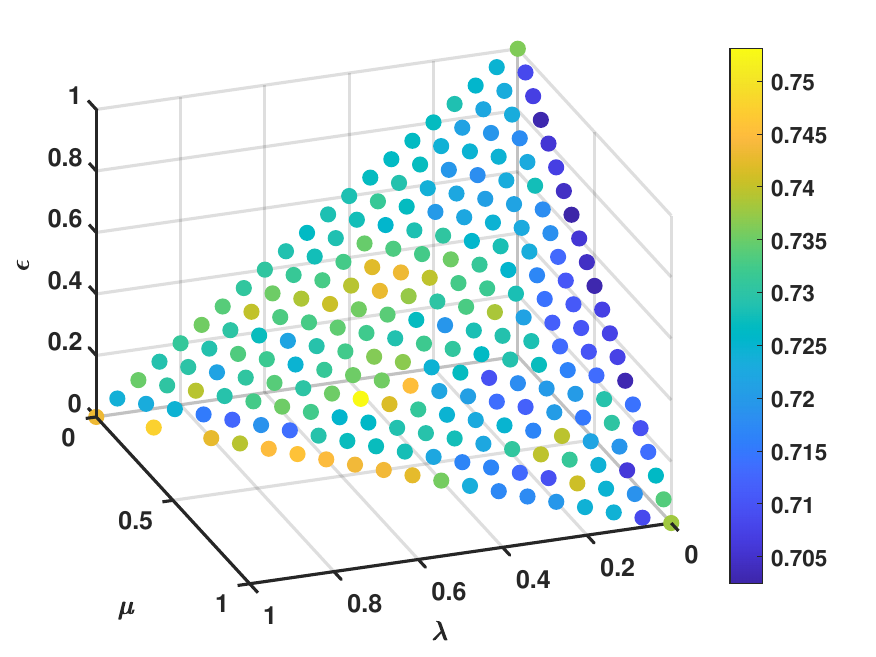}
\label{fig:6c}}
\hfil
\subfloat[K-L divergence]{\includegraphics[width=0.23\textwidth]{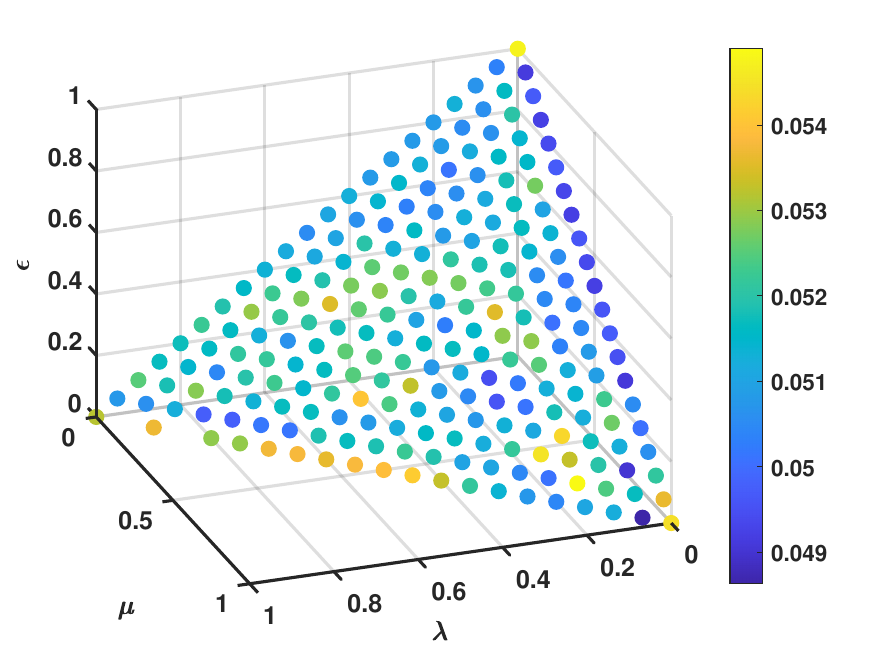}
\label{fig:6d}}\\
\subfloat[Cosine coefficient]{\includegraphics[width=0.23\textwidth]{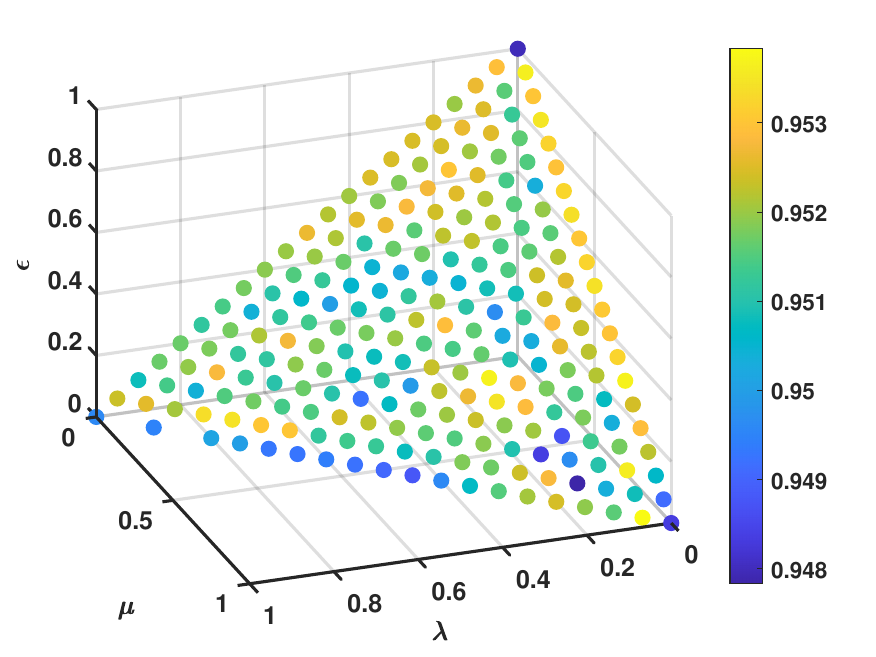}
\label{fig:6e}}
\hfil
\subfloat[Intersection similarity]{\includegraphics[width=0.23\textwidth]{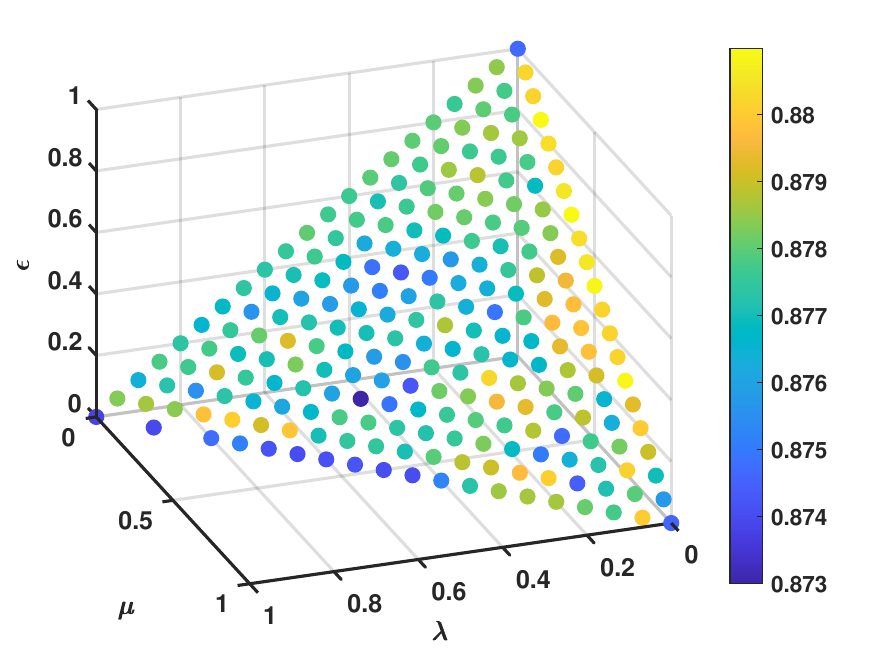}
\label{fig:6f}}
\caption{Parameter sensitivity on S-JAFFE}
\label{fig:6}
\end{figure}

Fig.~\ref{fig:6} presents the results of the parameter sensitivity analysis conducted on the S-JAFFE dataset. The parameters $\lambda$, $\mu$, and $\varepsilon$ are varied under the constraint $\lambda + \mu + \varepsilon = 1$, and their impacts are assessed using six evaluation metrics. As shown in Fig.~\ref{fig:6}, the performance exhibits considerable variation across different parameter combinations, indicating that the fusion weights ($\lambda$, $\mu$, and $\varepsilon$) have a significant influence on the final outcomes. Notably, for most metrics, optimal performance is achieved when the weights are relatively balanced, rather than heavily skewed towards a single component. This suggests that jointly integrating the three sources of information, i.e., the intrinsic relationships within individual clusters, the interactions across different clusters (capturing distance information), and the interactions across different clusters (capturing directional information), can effectively contribute to performance improvements. These findings validate the effectiveness of the adopted serial fusion scheme and offer practical insights for selecting suitable hyperparameter values.

\subsection{Comparison of Computational Complexity}

In Tab.~\ref{tab:10}, we present the computational complexity of several LDL algorithms, where $n$, $m$, and $p$ denote the number of instances, features, and labels, respectively. For LDL-SCL, $C$ is the cost of calculating the gradient. For LDL-LDM, $g$ is the number of clusters. For LDL-LIFT-SAP, both the base models and the meta model are trained using the SA-BFGS~\cite{geng:LDL}. Specifically, for each label $y_j$, the computational complexity of the corresponding base model is $\mathcal{O}(C_{base}^j) = \mathcal{O}\Big(T_j\big(n(2{m_j}^2 + {m_j^*}^2)+(2{m_j}^2 + {m_j^*}^2)^2\big)\Big)$, where $T_j$ denotes the number of iterations required for training the $j$-th base model. The computational complexity of the meta model is given by $\mathcal{O}(C_{meta}) = \mathcal{O}\big(T(np+p^2)\big)$, where $T$ is the number of iterations for meta model training. Overall, the computational complexity of LDL-LIFT-SAP is $\mathcal{O}\Big(\sum\limits_{j=1}^p T_j\big(n(2{m_j}^2 + {m_j^*}^2)+(2{m_j}^2 + {m_j^*}^2)^2 \big) + T(np+p^2)\Big)$.

\begin{table}[!t]
\setlength{\tabcolsep}{2pt}
\caption{Comparison of computational complexity.}
\centering
\begin{tabular}{ccccc}
\hline
Algorithms & Computational complexity  \\
\hline
EDL-LRL~\cite{jia-et-al:EDL-LRL} & $\mathcal{O}(m^2p^2+mnp+n^2p+p^3)$\\
LDL-LCLR~\cite{ren-et-al:LDL-LCLR} & $\mathcal{O}(m^2p^2 + mnp + n^2p + p^3)$\\
LDL-SCL~\cite{jia-et-al:LDL-SCL} & $\mathcal{O}(C)$\\
LDL-LDM~\cite{wang-geng:LDL-LDM} & $\mathcal{O}(np^3+gp^4+mnp)$\\
LIFT-SAP (ours) & $\mathcal{O}\Big(\sum\limits_{j=1}^p \big(n^2(2m_j+m_j^*) + n(2m_j^2+{m_j^*}^2)m\big)\Big)$\\
LDL-LIFT-SAP (ours) & 
\makecell[l]{$\mathcal{O}\Big(\sum\limits_{j=1}^p T_j\big(n(2{m_j}^2 + {m_j^*}^2)+(2{m_j}^2 + {m_j^*}^2)^2 \big)$\\
\qquad $+ T(np+p^2)\Big)$} \\
\hline
\end{tabular}
\label{tab:10}
\end{table}

\section{Conclusion}
\label{sec:conclusion}
In this paper, we propose a novel label-specific feature (LSF) construction strategy, LIFT-SAP, which extends the widely used LIFT framework by introducing structural anchor points (SAPs) to capture inter-cluster interactions that are typically ignored by existing prototype-based LIFT approaches. Unlike traditional LSFs constructed solely based on Euclidean distances, LIFT-SAP incorporates both distance and directional information relative to SAPs, resulting in a more robust and comprehensive representation of each instance. Building upon this enriched feature space, we further develop an LDL algorithm, LDL-LIFT-SAP, which integrates multiple predictions derived from diverse LSF spaces into a unified label distribution, thereby addressing label ambiguity more effectively.

Extensive experiments on $15$ real-world LDL datasets demonstrate the superiority of the proposed approach. LIFT-SAP consistently outperforms the widely used LIFT in constructing discriminative features, while LDL-LIFT-SAP achieves significant improvements in predictive performance compared to seven well-established LDL algorithms across six evaluation metrics. These results confirm the effectiveness of modeling inter-cluster relationships and highlight the advantage of leveraging multi-perspective feature construction.

Despite the impressive performance of LIFT-SAP and LDL-LIFT-SAP, we acknowledge some limitations, which motivate further investigations in this direction:

(1) SAPs are constructed as midpoints between cluster centers, which generally works well but may produce redundant or less informative anchors when clusters are poorly separated. Future work could explore adaptive SAP selection, such as graph-based modeling or prototype pruning/refinement, to enhance anchor quality.

(2) While simple serial fusion of LSFs generated by SAPs into LIFT proves effective, future work could investigate more fusion strategies, such as joint feature learning or learnable fusion coefficients, to better exploit the complementary strengths of different feature types.

(3) Although LIFT-SAP has shown satisfactory performance in LDL tasks, it is not limited to this setting. Future work could explore its extension to MLL and PLL tasks, further enhancing its applicability across a broader range of label-ambiguous learning scenarios. 

\section*{Acknowledgments}
The authors would like to thank the anonymous reviewers and the editor for their constructive and valuable comments.

\bibliographystyle{IEEEtran}

\bibliography{ref}

\vfill

\end{document}


\title{Supplementary Materials:\\ From Distance to Direction: Structure-aware Label-specific\\ Feature Fusion for Label Distribution Learning}

\author{Suping Xu, Chuyi Dai, Lin Shang,~\IEEEmembership{Member,~IEEE,} \\Changbin Shao, Xibei Yang, and Witold Pedrycz,~\IEEEmembership{Life Fellow,~IEEE}

\thanks{S. Xu and C. Dai are with the Department of Electrical and Computer Engineering, University of Alberta, Edmonton, AB T6G 2R3, Canada (E-mail: supingxu@yahoo.com; suping2@ualberta.ca, cdai4@ualberta.ca).}
\thanks{L. Shang is with the School of Computer Science, Nanjing University, Nanjing 210023, China, and also with the State Key Laboratory for Novel Software Technology, Nanjing University, Nanjing 210023, China (E-mail: shanglin@nju.edu.cn).}
\thanks{C. Shao and X. Yang are with the School of Computer, Jiangsu University of Science and Technology, Zhenjiang 212003, China (E-mail: shaocb@just.edu.cn, jsjxy\_yxb@just.edu.cn).}
\thanks{W. Pedrycz is with the Department of Electrical and Computer Engineering, University of Alberta, Edmonton, AB T6G 2R3, Canada, also with the Institute of Systems Engineering, Macau University of Science and Technology, Taipa 999078, China, and also with the Research Center of Performance and Productivity Analysis, Istinye University, Istanbul 34396, Türkiye (E-mail: wpedrycz@ualberta.ca).}

\thanks{Manuscript received X X, 2025; revised X X, 2025.}}

\markboth{Journal of \LaTeX\ Class Files,~Vol.~X, No.~X, August~2025}%
{Shell \MakeLowercase{\textit{et al.}}: A Sample Article Using IEEEtran.cls for IEEE Journals}

\maketitle

\section{Experimental Supplement to Ablation Analysis}

\begin{figure}[htbp]
\centering
\includegraphics[width=0.42\textwidth]{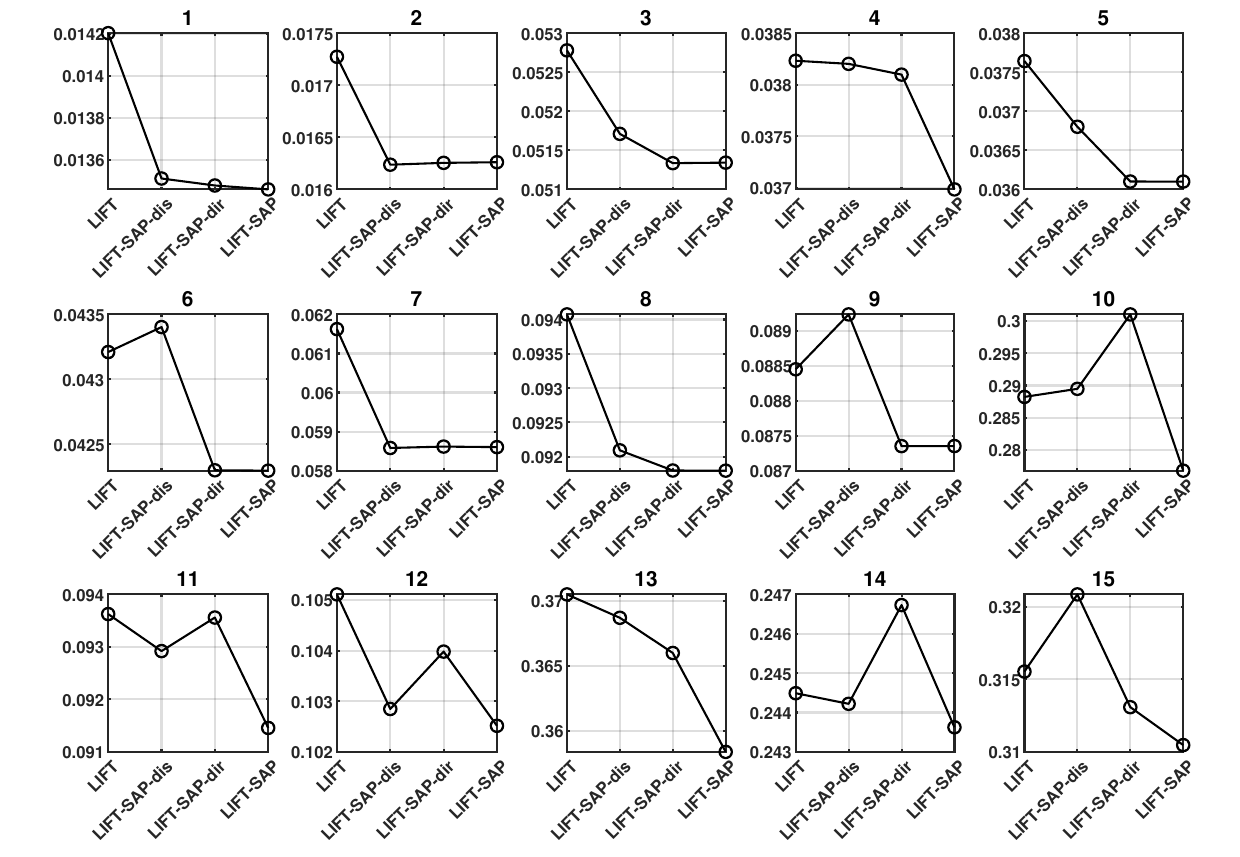}
\caption{Ablation analysis on the $15$ LDL datasets (\emph{Chebyshev distance} $\downarrow$).}
\label{fig:s1}
\end{figure}

\begin{figure}[htbp]
\centering
\includegraphics[width=0.42\textwidth]{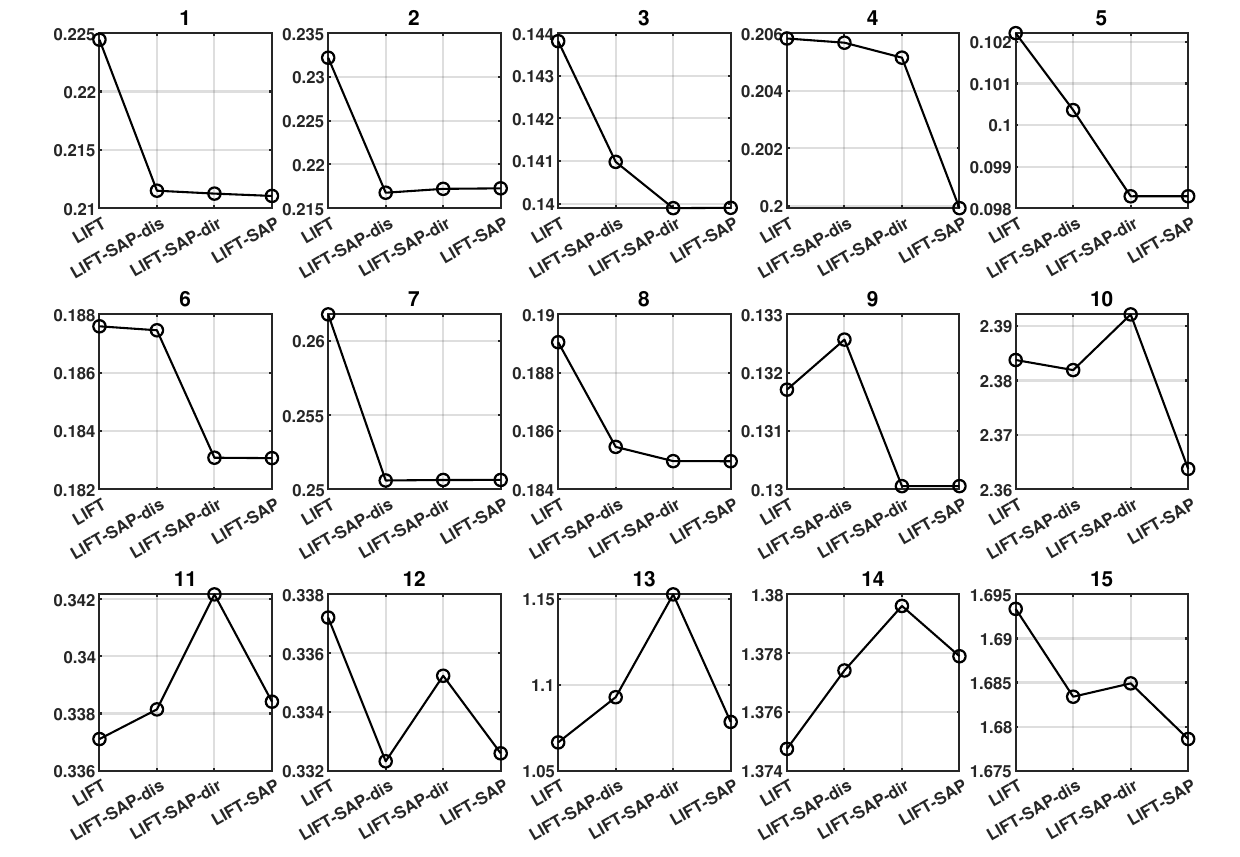}
\caption{Ablation analysis on the $15$ LDL datasets (\emph{Clark distance} $\downarrow$).}
\label{fig:s2}
\end{figure}

\begin{figure}[htbp]
\centering
\includegraphics[width=0.42\textwidth]{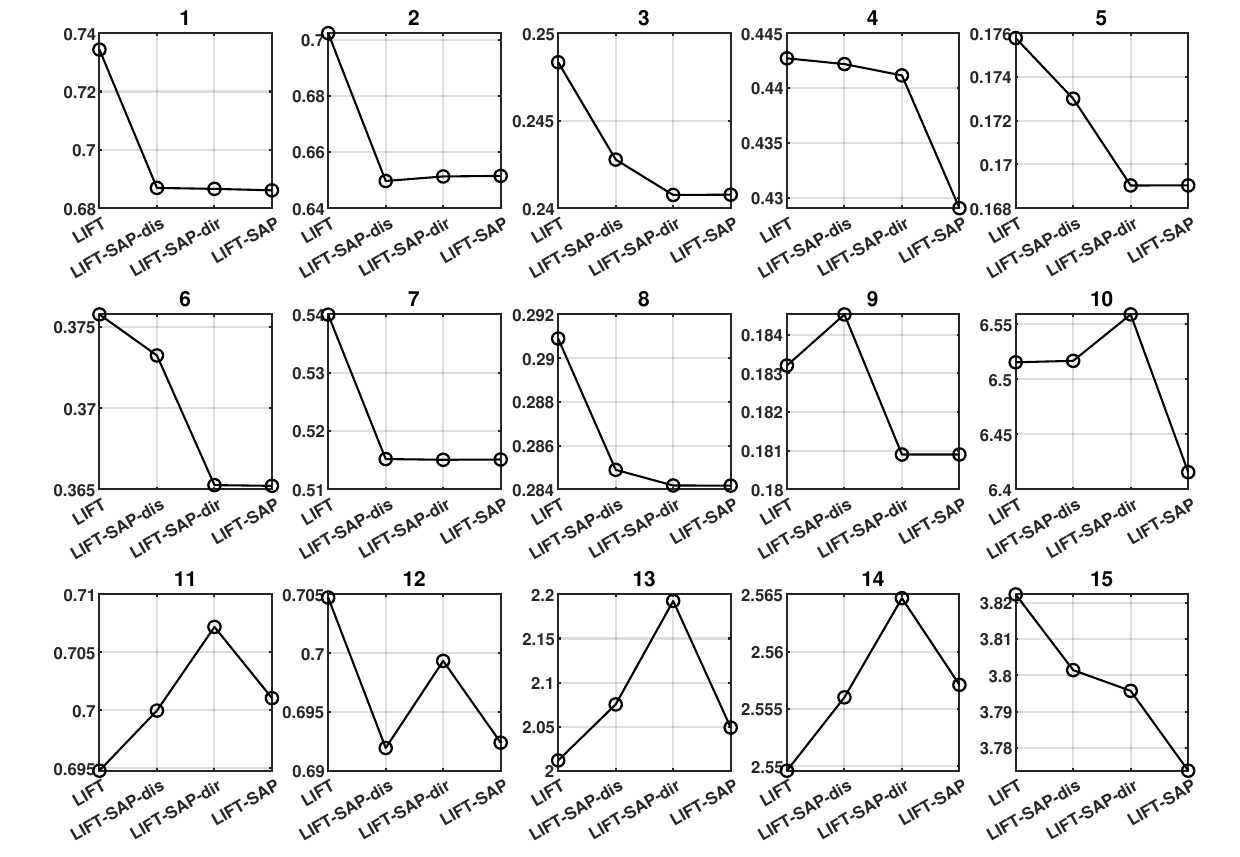}
\caption{Ablation analysis on the $15$ LDL datasets (\emph{Canberra metric} $\downarrow$).}
\label{fig:s3}
\end{figure}

\begin{figure}[htbp]
\centering
\includegraphics[width=0.42\textwidth]{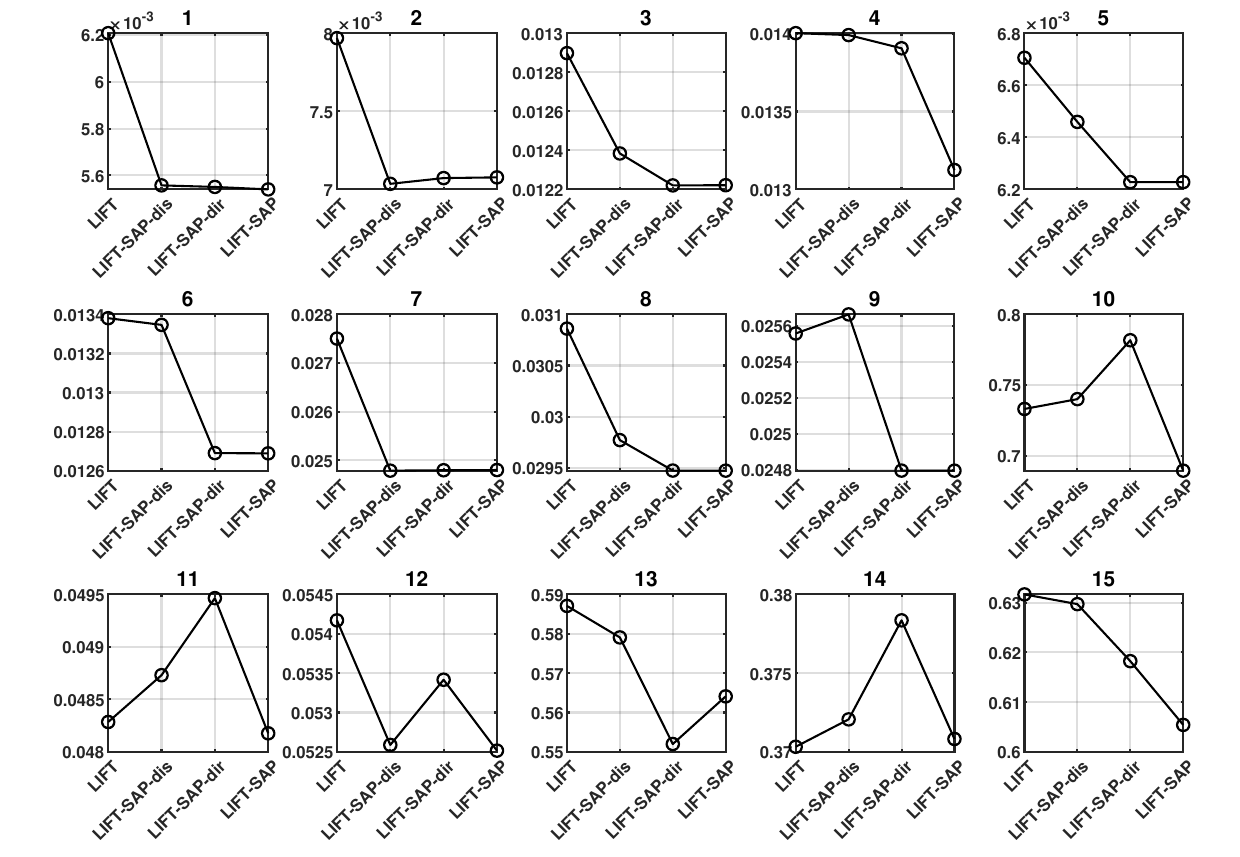}
\caption{Ablation analysis on the $15$ LDL datasets (\emph{K-L divergence} $\downarrow$).}
\label{fig:s4}
\end{figure}

\begin{figure}[htbp]
\centering
\includegraphics[width=0.42\textwidth]{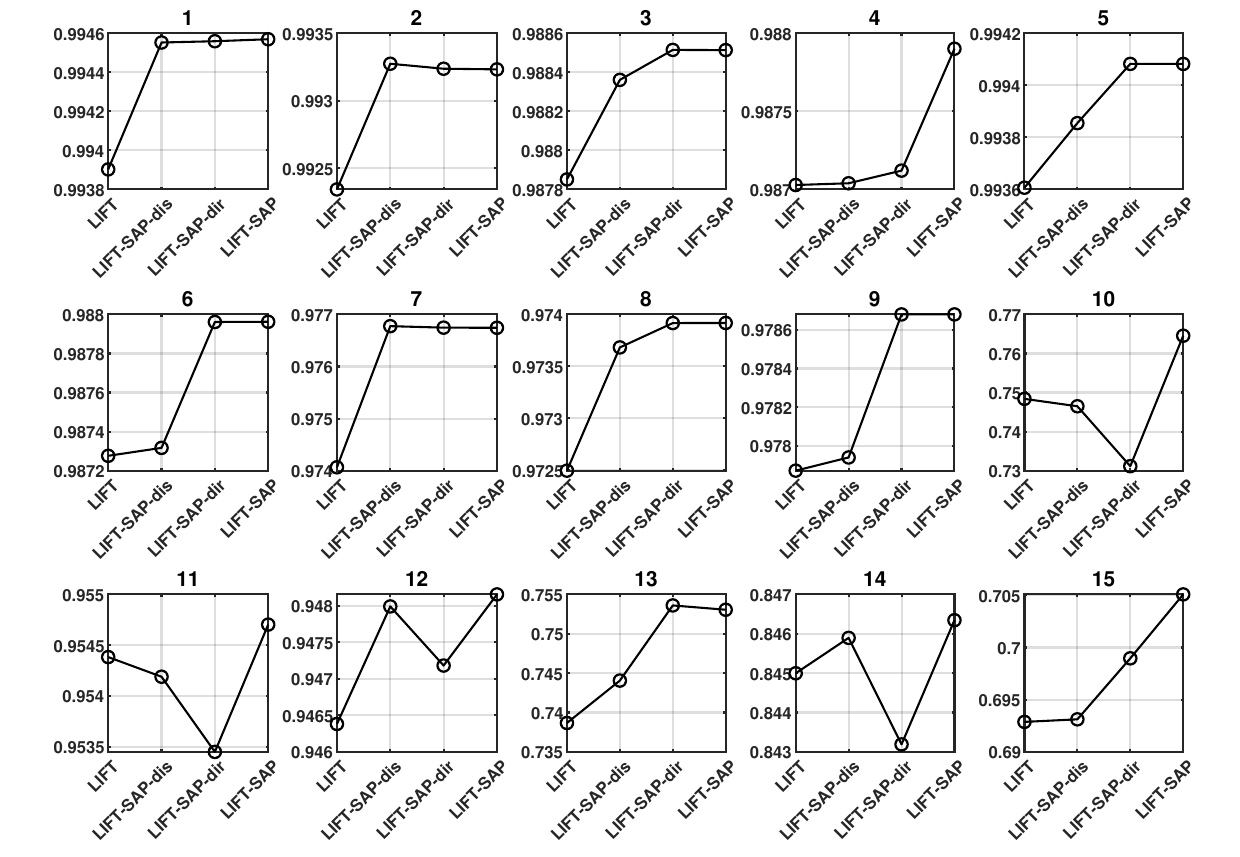}
\caption{Ablation analysis on the $15$ LDL datasets (\emph{Cosine coefficient} $\uparrow$).}
\label{fig:s5}
\end{figure}

\begin{figure}[H]
\centering
\includegraphics[width=0.42\textwidth]{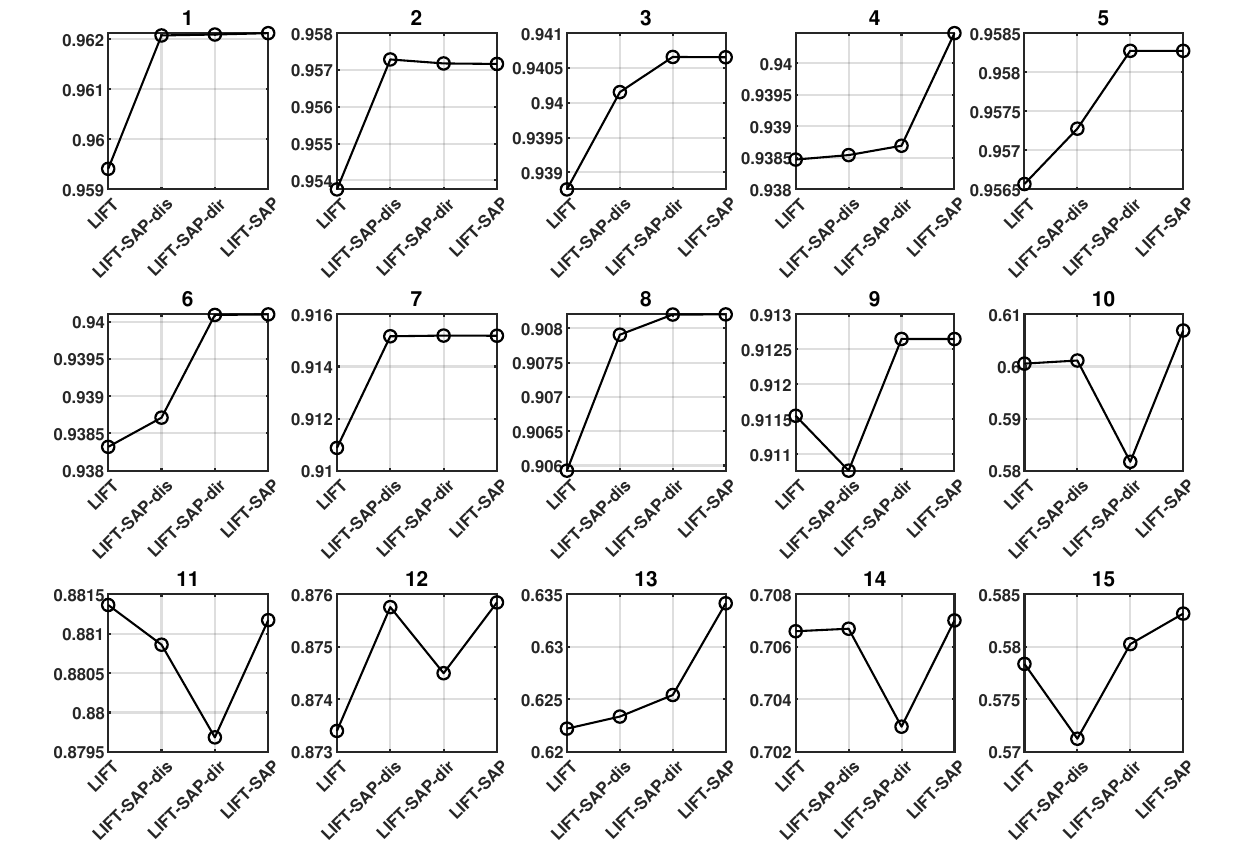}
\caption{Ablation analysis on the $15$ LDL datasets (\emph{Intersection similarity} $\uparrow$).}
\label{fig:s6}
\end{figure}

We reported the performance of LIFT-SAP and its three variants, namely, LIFT, LIFT-SAP without direction relative to SAPs, LIFT-SAP without distance relative to SAPs, and LIFT-SAP, on four datasets: \emph{Yeast-spo5} (Bioinformatics), \emph{Natural Scene} (Natural scene recognition), \emph{M$^2$B} (Facial beauty assessment), and \emph{Emotions6} (Facial expression recognition). Here, we extend this by presenting the ablation analysis across all $15$ LDL datasets using six evaluation metrics. (Figs.\ref{fig:s1}-\ref{fig:s6}). We can draw similar conclusions.

